\pdfoutput=1

\documentclass[11pt]{article}

\usepackage[table]{xcolor}

\usepackage{acl}

\usepackage{acl}
\usepackage{float}
\usepackage{hyperref}
\usepackage{amsmath} 
\usepackage{amssymb} 
\usepackage{cleveref}
\usepackage{graphicx} 
\usepackage{natbib} 
\usepackage{booktabs} 
\usepackage{array}    %

\usepackage{times}
\usepackage{latexsym}
\usepackage{booktabs}
 \usepackage{amsmath} 
\usepackage[skins]{tcolorbox} 
\usepackage{enumitem}    
\usepackage{geometry}  

\usepackage{multirow}
\usepackage{amssymb}
\usepackage{pifont}
\usepackage{xspace}

\definecolor{bblue}{HTML}{4F81BD}
\definecolor{rred}{HTML}{C0504D}
\definecolor{ggreen}{HTML}{9BBB59}
\usepackage{times}
\usepackage{latexsym}

\usepackage[T1]{fontenc}

\usepackage[utf8]{inputenc}

\usepackage{microtype}

\usepackage{inconsolata}

\usepackage{graphicx}
\usepackage{cleveref}
\usepackage[utf8]{inputenc}
\usepackage{csquotes}

\usepackage{times}
\usepackage{latexsym}
\usepackage{booktabs}
\usepackage{microtype}
\usepackage{amssymb}
\usepackage{pifont}
\usepackage{graphicx}
\usepackage{placeins}
\usepackage{natbib}
\usepackage{longtable}
\usepackage{cleveref}
\usepackage{enumitem}
\usepackage{booktabs}
\usepackage{pdfpages}

\usepackage{tabularx}
\usepackage{multirow}
\usepackage{xspace}

\usepackage{linguex}
\alignSubExtrue

\usepackage[verbose]{newunicodechar}

\newcolumntype{T}{>{\tiny}l}

\usepackage{rotating}
\usepackage{times}
\usepackage{latexsym}
\usepackage{microtype}
\usepackage{amssymb}
\usepackage{pifont}
\usepackage{graphicx}
\usepackage{placeins}
\usepackage{natbib}
\usepackage{tablefootnote}
\usepackage{longtable}
\usepackage{cleveref}
\usepackage{enumitem}
\usepackage{booktabs}
\usepackage{pdfpages}
\usepackage[T1]{fontenc}

\usepackage{tabularx}
\usepackage{multirow}
\usepackage{xspace}
\usepackage{times}
\usepackage{latexsym}
\usepackage[T1]{fontenc}
\usepackage[utf8]{inputenc}
\usepackage{microtype}
\usepackage{inconsolata}
\usepackage{graphicx}
\usepackage{booktabs}
\usepackage{longtable}
\usepackage{lscape}
\usepackage{array}
\usepackage{booktabs}
\usepackage{siunitx}
\usepackage{adjustbox}

\usepackage[textsize=footnotesize]{todonotes}
\usepackage{linguex}
\usepackage{subcaption}
\alignSubExtrue
\usepackage[verbose]{newunicodechar}

\usepackage{inconsolata}

\usepackage{makecell}
\global

\tcbset{
  aibox/.style={
    width=474.18663pt,
    top=10pt,
    colback=white,
    colframe=black,
    colbacktitle=black,
    enhanced,
    center,
    attach boxed title to top left={yshift=-0.1in,xshift=0.15in},
    boxed title style={boxrule=0pt,colframe=white,},
  }
}
\newtcolorbox{AIbox}[2][]{aibox,title=#2,#1}
\newcommand{\bestmono}{\cellcolor{blue!30}}

\newcommand{\bestllm}{\cellcolor{orange!30}}
\definecolor{blue1}{RGB}{234, 230, 255} 
\definecolor{blue2}{RGB}{194, 200, 255} 
\definecolor{blue3}{RGB}{154, 170, 255}

\definecolor{red1}{RGB}{255,245,238}
\definecolor{red2}{RGB}{255,228,225}
\definecolor{red3}{RGB}{255,188,185}

\definecolor{light-gray}{HTML}{E5E4E2}
\definecolor{light-cyan}{HTML}{E0FFFF}
\newcolumntype{R}{>{\raggedleft\arraybackslash}p{0.28cm}}
\newcolumntype{L}{>{\raggedleft\arraybackslash}p{0.23cm}}

\usepackage{microtype}

\newcommand\ours{\textbf{\textsc{BLE}n\textsc{D}}}

\title{SemEval-2026 Task 7: Everyday Knowledge Across Diverse Languages and Cultures}

\author{
Nedjma Ousidhoum$^{1}$, Junho Myung$^{2}$, Carla Perez-Almendros$^{1}$, Jiho Jin$^{2}$,\\
\textbf{Amr Keleg$^{3}$, Meriem Beloucif$^{4}$, Yi Zhou$^{1}$, Rodrigo Agerri$^{5}$, Vladimir Araujo$^{6}$, Naomi Baes$^{7}$,} \\
\textbf{James Barry$^{8}$, Joanne Boisson$^{1}$, Nancy F. Chen$^{9}$, Christine de Kock$^{7}$, Aleksandra Edwards$^{1}$,}\\
\textbf{Joseba Fernandez de Landa$^{5}$, Mohamed Fazli Imam$^{3}$, Huda Hakami$^{10}$, Shu-Kai Hsieh$^{11}$,}\\
\textbf{Joseph Marvin Imperial$^{12,13}$, Roy Ka-Wei Lee$^{14}$, Zhengyuan Liu$^{9}$, Chenyang Lyu$^{15}$,}\\
\textbf{Younes Samih$^{8}$, Johan Sjons$^{4}$, Bryan Tan$^{14}$, Asahi Ushio$^{16}$\thanks{Work done while the author was at Cardiff University.}, Weihua Zheng$^{14}$,}\\
\textbf{Alice Oh$^{2}$, Jose Camacho-Collados$^{1}$}\\
\footnotesize{
$^{1}$Cardiff University, $^{2}$KAIST, $^{3}$MBZUAI, $^{4}$Uppsala University, $^{5}$HiTZ Center, University of the Basque Country EHU, $^{6}$Sailplane AI,}\\
\footnotesize{$^{7}$University of Melbourne, $^{8}$IBM Research, $^{9}$Agency for Science, Technology and Research (A*STAR), Singapore,}\\
\footnotesize{$^{10}$Taif University, $^{11}$National Taiwan University, $^{12}$National University Philippines, $^{13}$University of Bath, } \\
\footnotesize{$^{14}$Singapore University of Technology and Design, $^{15}$Alibaba, $^{16}$Google}
}

\begin{document}
\maketitle
\begin{abstract}
We present our shared task on evaluating the adaptability of LLMs and NLP systems across multiple languages and cultures. The task data consist of an extended version of our manually constructed \ours{} benchmark \cite{myung2024blend}, covering more than 30 language--culture pairs, predominantly representing low-resource languages spoken across multiple continents. As the task is designed strictly for evaluation, participants were not permitted to use the data for training, fine-tuning, few-shot learning, or any other form of model modification.
Our task includes two tracks: (a) Short-Answer Questions (SAQ) and (b) Multiple-Choice Questions (MCQ). Participants were required to predict labels and were allowed to submit any NLP system and adopt diverse modelling strategies, provided that the benchmark was used solely for evaluation. The task attracted more than 140 registered participants, and we received final submissions from 62 teams, along with 19 system description papers.
We report the results and present an analysis of the best-performing systems and the most commonly adopted approaches. Furthermore, we discuss shared insights into open questions and challenges related to evaluation, misalignment, and methodological perspectives on model behaviour in low-resource languages and for under-represented cultures.\footnote{Our data and resources are available at \url{https://github.com/BLEnD-SemEval2026/SemEval-2026-Task-7}.}

\end{abstract}

\section{Introduction}
\begin{figure}
    \centering
    \includegraphics[width=0.99\linewidth]{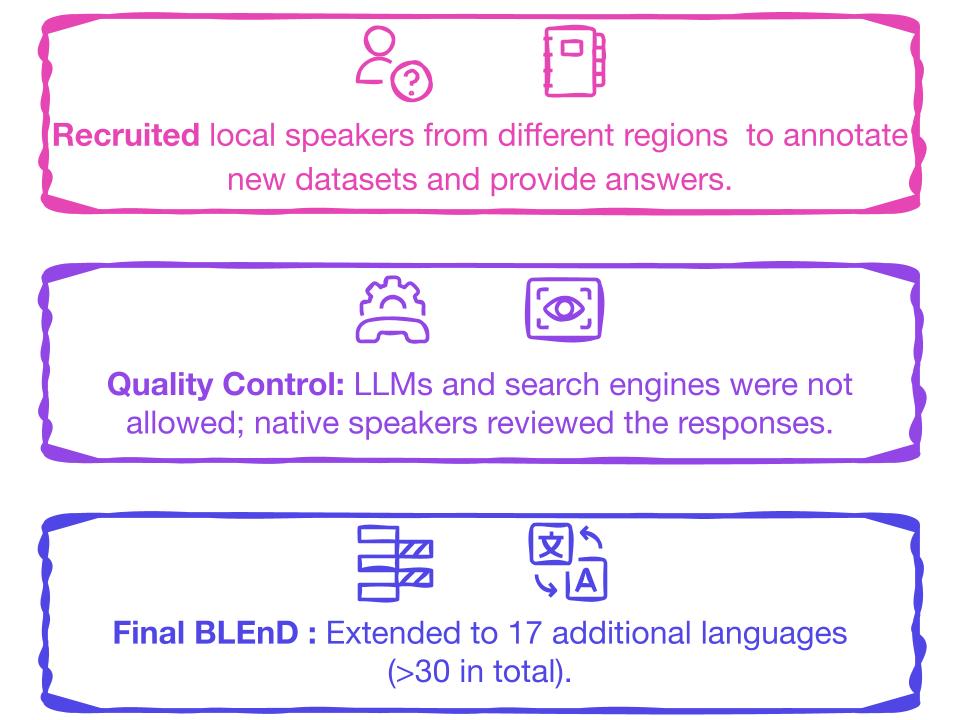}
    \caption{\textbf{Data creation pipeline:} recruitment of local speakers for annotation, native-speaker quality control without the use of LLMs or search engines, and extension to 17 additional languages.} 
    \label{fig:pipeline}
\end{figure}

\begin{figure}[ht]
    \centering
    \includegraphics[width=0.99\linewidth]{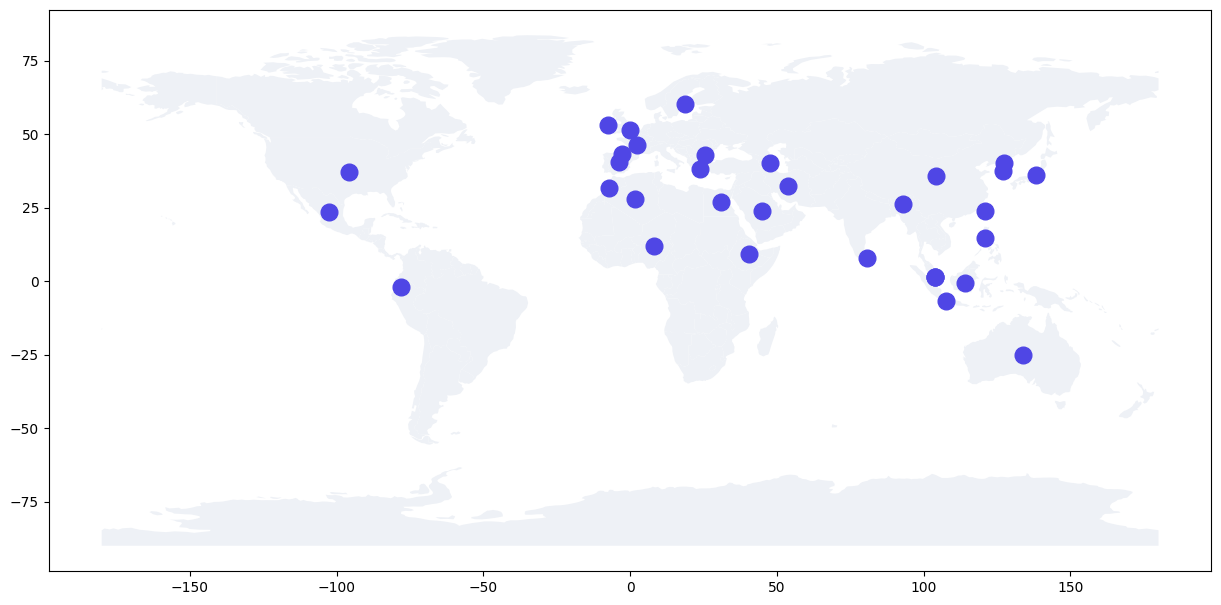}
    \caption{\textbf{Language--culture pairs represented in our \ours{} benchmark.} 
    Africa: Arabic (Algeria, Egypt, Morocco), Amharic (Ethiopia), Hausa (Northern Nigeria). 
    Asia: Assamese (Assam, India), Azerbaijani (Azerbaijan), Mandarin (China), Indonesian (Indonesia), Javanese (West Java, Indonesia), Persian (Iran), Korean (North and South Korea), Arabic (Saudi Arabia), Japanese (Japan), Tagalog (Philippines), Tamil (Sri Lanka, Singapore), Taiwanese Mandarin (Taiwan), Singaporean Mandarin and Malay (Singapore). 
    Australia: English (Australia). 
    Europe: Greek (Greece), Spanish (Spain), English (UK), French (France), Bulgarian (Bulgaria), Swedish (Sweden), Irish (Ireland), Basque (Basque Country).}
    \label{fig:blend_languages}
    \vspace{5pt}
\end{figure}

The global deployment of LLMs and NLP systems necessitates cultural adaptability in both multilingual and English settings~\cite{adilazuarda-etal-2024-towards, pawar2024surveyculturalawarenesslanguage, liu2025culturallyawareadaptednlp}. However, such models often exhibit substantial limitations in culture-specific knowledge, particularly when handling under-resourced languages or non-Western regions. They tend to generate responses that reflect Western-centric perspectives~\cite{durmus2023towards, koto2024indoculture, naous-etal-2024-beer} or reproduce stereotypes present in their training data~\cite{navigli2023biases, zhou-etal-2023-predictive, kaneko2024evaluating, zhou-etal-2024-evaluating-short}.
Existing benchmarks~\cite{kim2024click, son2024hae, fung2024massively, koto2024indoculture} predominantly rely on monolingual datasets or online resources such as Wikipedia, which often fail to capture the nuanced realities of everyday life across diverse cultural contexts. To address this limitation, we extend our previously developed \ours{} benchmark~\cite{myung2024blend}, as shown in \autoref{fig:pipeline}, to 33 language--culture pairs listed in \autoref{fig:blend_languages}. Specifically, the original \ours{} includes: Amharic (Ethiopia, am-ET), Arabic (Algeria, ar-DZ), Assamese (Assam, India, as-AS), Azerbaijani (Azerbaijan, az-AZ), Greek (Greece, el-GR), English (United Kingdom, en-GB; USA, en-US), Spanish (Spain, es-ES; Mexico, es-MX), Persian (Iran, fa-IR), Hausa (Northern Nigeria, ha-NG), Indonesian (Indonesia, id-ID), Korean (North Korea, ko-KP; South Korea, ko-KR), Javanese (West Java, Indonesia, su-JB), and Mandarin (China, zh-CN).
For this shared task, we added extensions for Arabic (Egypt, ar-EG; Morocco, ar-MA; Saudi Arabia, ar-SA), Bulgarian (Bulgaria, bg-BG), English (Australia, en-AU; Singapore, en-SG), Spanish (Ecuador, es-EC), Basque (Basque Country, eu-PV), French (France, fr-FR), Irish (Ireland, ga-IE), Japanese (Japan, ja-JP), Swedish (Sweden, sv-SE), Tamil (Sri Lanka, ta-LK; Singapore, ta-SG), Tagalog (Philippines, tl-PH), Singaporean Mandarin (Singapore, zh-SG), Taiwanese Mandarin (Taiwan, zh-TW), and Malay (Singapore, ms-SG).

\begin{figure}[!h]
\centering
\includegraphics[width=0.99\linewidth]{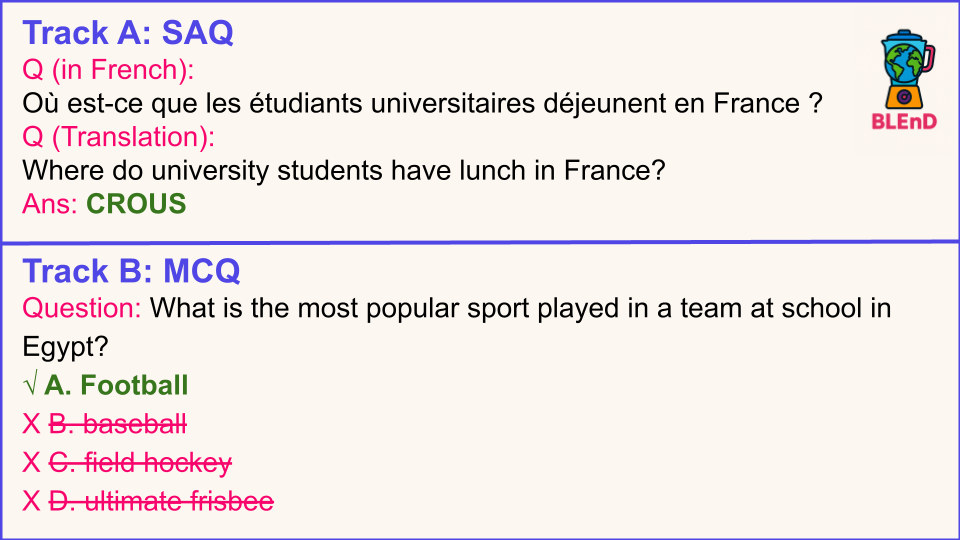}
\caption{Examples of a short answer question (Track A) and a multiple-choice question (Track B) from our \ours{} dataset.} 
\label{fig:blend_examples}
\vspace{5pt}
\end{figure}

As our objective is to evaluate cultural adaptability across languages, \ours{} was used by participants exclusively for validation and testing and was withheld from system training to ensure that the results genuinely reflect a model's ability to generalise to unseen and culturally diverse contexts. Our task includes two tracks: (a)\ Short-Answer Questions (SAQ), and (b)\ Multiple-Choice Questions (see examples in \autoref{fig:blend_examples}). Participants could submit answers and compete in one or both tracks and in any language(s) of their choice. Our task attracted over 140 participants, with 62 final submissions and 19 teams submitting system description papers. Track~B (MCQ) received the highest number of submissions, followed by Track~A (SAQ), with most teams participating in multiple languages. We discuss the results and adopted approaches, as well as additional qualitative insights into the evaluation process, model behaviour, and design choices when tackling low-resource languages.

\section{Data}

We extend the \ours{} benchmark \cite{myung2024blend} by involving language leads as core team members, as shown in \autoref{fig:pipeline}. Each language lead recruited at least five native speakers to provide answers to language-specific questions. Below, we describe the process used to construct the newly added datasets.

\subsection{Data Collection}
\ours{} data instances consist of question--answer pairs. Answers are either short free-text responses (Track~A) or multiple-choice options (Track~B).

\paragraph{Initial \ours{} dataset construction}
To build the original \ours{} dataset, we recruited native speakers to create culturally grounded questions and answers across specific topics (food, sport, work life, family, education, and traditional holidays, celebrations, and leisure).
Annotators were instructed to:
\begin{itemize}[noitemsep,nolistsep]
\item write questions reflecting their cultural or country-specific background, or requiring common sense within that cultural context (e.g., \textit{When do people from [region X] typically have dinner? Please provide a time in HH:MM format, e.g., 20:00.});
\item avoid purely factual or yes/no questions, and ensure diversity in question types (e.g., \textit{Do people celebrate Chinese New Year in [region X]?});
\item cover a broad range of subtopics;
\item avoid stereotypical questions (e.g., \textit{Who usually does the house chores in [region X]?}), as these would be removed if identified later.
\end{itemize}
Answers had to be concrete and specific, using precise concepts, named entities, or time references where relevant. As some questions required cultural knowledge or common sense, there were cases with no objectively correct or incorrect responses, and multiple answers were acceptable. That is, each annotator could provide up to three answers, which were required to reflect the annotator's own cultural background. All questions and answers were written independently. Annotators were instructed not to use LLMs or search engines. Except for US and GB English, questions were translated into the respective local languages by native speakers. This process resulted in a dataset of 15{,}000 short-answer questions.

\paragraph{Dataset extension}
The extended benchmark covers 33 datasets curated and annotated by native speakers. Each lead for the additional 17 language--culture pairs recruited five local speakers who had spent at least half of their lives in that region. All teams followed the same guidelines to ensure consistency and comparability across languages and cultural contexts.
The extension to 17 additional languages was carried out by translating the English versions of the original \ours{} questions into the target languages (e.g., Tagalog, Japanese), except for Australian English, Ecuadorian Spanish, and Arabic variants (Egyptian, Moroccan, and Saudi). In these cases, the original English, Spanish, or Modern Standard Arabic questions were adapted to the respective country contexts. For example, ``What is the most common snack for school children in \textit{the U.S.}?'' was adapted by replacing \textit{the U.S.} with \textit{Australia}.

\paragraph{Answer collection}
Annotators were required to provide short answers to the questions based on their own cultural perspective. All responses had to be written in their native language and reflect their cultural or country-specific background.
Answers were required to be brief and concrete. Annotators were encouraged to use precise concepts, named entities, and time references where appropriate. There were no objectively correct or incorrect answers. Each question had to receive at least one primary answer; additional answers could be provided where appropriate. All answers had to be produced independently, without consulting LLMs or search engines. Annotators could select ``no answer'', ``idk'' (I do not know), or ``does not apply to my culture'' if the question was not applicable (e.g., a question about the Mid-Autumn Festival posed to a Bulgarian annotator).

\subsection{Quality Control}
\subsubsection{Annotator recruitment}
A pre-test consisting of 30 questions was used for qualification, and attention-check questions were included throughout the annotation task. For high-resource languages with strong representation on Prolific (Japanese, French, and Australian English), we used the platform to collect answers, and participants were required to provide demographic information to verify that they met the study criteria. By contrast, for low-resource languages, language leads recruited annotators directly, as native speakers of these languages are under-represented on Prolific.
\subsubsection{Aggregation}
Language leads assessed answers based on whether they addressed the question. For quality control:
\begin{itemize}[nolistsep, noitemsep]
\item Responses were aggregated by merging identical or similar answers into a single category (e.g.\ ``apple'' and ``apples''). Specifically:
\begin{itemize}[nolistsep, noitemsep]
\item minor spelling, wording, grammatical, or tonal variations were normalised;
\item hierarchical categories (e.g.\ ``fruit'' vs.\ ``mango'') were treated as distinct concepts and assigned to different groups.
\end{itemize}
\item For questions with no answer or those that were inapplicable to specific language--culture pairs, responses were retained as such.
\item Answers that did not appropriately address the question (e.g., the question asks about a fruit and the annotator answers ``noodles'') were discarded.
\end{itemize}

\paragraph{English translation}
Answers were translated into English using a translation tool, then manually reviewed by a native speaker and corrected where necessary. Where relevant, multiple English translations or notational variants were provided.

\paragraph{Synonyms}
Annotators listed alternative forms in their language referring to the same concept (i.e., synonyms), separated by line breaks. For example, if ``durén'' appeared, alternative forms such as ``duren'' (without tone marks) or ``buah duren'' (\textit{durian fruit}) were listed. If a synonym already appeared among the answers for that question with the same group number, it was not listed again.

\subsubsection{MCQ verification}
Annotators received automatically generated files containing multiple-choice questions and answer options. They reviewed each question--answer pair to ensure that the designated correct answer was the only correct option. If a question had more than one correct answer, it was flagged for removal.
If all answer options were incorrect, annotators could add a comment or propose a more appropriate answer. If alternative options were clearly better than the proposed ones, improvements could be suggested in the comments.

\section{Task Description}
\begin{table*}[t]
\centering
\scriptsize

\begin{subtable}{\textwidth}
\centering
\setlength{\tabcolsep}{3pt}
\begin{adjustbox}{width=\textwidth}
\begin{tabular}{l*{16}{S}}
\toprule
\textbf{Team} &
\texttt{ar-EG} & \texttt{en-EG} & \texttt{ar-MA} & \texttt{en-MA} &
\texttt{ar-SA} & \texttt{en-SA} & \texttt{eu-PV} & \texttt{en-PV} &
\texttt{bg-BG} & \texttt{en-BG} & \texttt{zh-SG} & \texttt{en-AU} &
\texttt{fr-FR} & \texttt{en-FR} & \texttt{ga-IE} & \texttt{en-IE} \\
\midrule
\textbf{1. king001} & \bestmono60.4 & \bestmono62.2 & 32 & \bestmono47.8 & \bestmono56.4 & \bestmono57.4 & \bestmono56.4 & 51.6 & \bestmono57.8 & \bestmono57 & \bestmono78 & \bestmono71.4 & 72.6 & \bestmono64 & 48.8 & \bestmono59.4 \\
\rowcolor{gray!25}
\textbf{\texttt{GPT4.1}} & \bestllm64.6 & 58.2 & 31.2 & 43.4 & \bestllm61.4 & 54.4 & 46 & \bestllm53.6 & \bestllm60.2 & \bestllm57 & \bestllm80.8 & \bestllm72 & \bestllm78.6 & \bestllm67.4 & 41 & \bestllm60 \\
\textbf{2. K-NLPers} & 53.2 & 52.8 & \bestmono36.4 & 36.4 & 46.8 & 46.6 & 44 & 42 & 48.4 & 47.6 & 74.4 & 68 & 71.6 & 58.2 & \bestmono54.8 & 54.4 \\
\textbf{3. GUIR} & 53 & 49.6 & 24 & 38.8 & 53.4 & 50.6 & 48.4 & 46 & 58 & 49.8 & 72.2 & 66.8 & \bestmono75.2 & 61 & 39.2 & 53.6 \\
\rowcolor{gray!25}
\textbf{\texttt{Qwen3-2}} & 56 & 52.6 & 28.8 & 36 & 50.4 & 49.2 & 31.4 & 47.6 & 50 & 48.6 & 73.8 & 66 & 71.6 & 59.8 & 24.2 & 51.6 \\
\textbf{4. LocuPrompt} & 48 & 59.6 & 13.4 & 24.4 & 40 & 50.8 & 54.6 & \bestmono53 & 34.6 & 35 & 64.6 & 70.2 & 64.8 & 59.2 & 48.8 & 52 \\
\textbf{5. wangkongqiang} & 44.6 & 52.2 & 19.2 & 35.8 & 49.2 & 48.8 & 43.2 & 40.2 & 48 & 50 & 77.6 & 68 & 70.6 & 56.4 & 32.8 & 56.6 \\
\textbf{6. FLANS} & 22.6 & 34.4 & 10 & 25.6 & 21.4 & 34 & 0 & 21.8 & 26 & 32.8 & 0.2 & 36.4 & 36 & 40 & 9.4 & 41.4 \\
\textbf{7. Agentic} & 26 & {-} & 9.8 & {-} & {-} & {-} & 3.8 & {-} & 21.8 & {-} & 50.2 & 48.6 & 44.4 & {-} & 9.2 & {-} \\
qinchihongye & 57.4 & 57.8 & 30.6 & 47.6 & 57.2 & 56 & 58.2 & 50.4 & 59.2 & 55.2 & 72.8 & 68.2 & 73.2 & 61.2 & 46.2 & 59.2 \\
rustycoder & 48.2 & 49.6 & 28.4 & 35.6 & 49.4 & 48.6 & 33.4 & 40.6 & 47.2 & 49.6 & 77.4 & 70.2 & 65.8 & 54 & 45.8 & 51.8 \\
fedikallel & 40 & 45 & 22.8 & 32.2 & 45.6 & 40.6 & 43 & 37.2 & 53.4 & 45.4 & 68.6 & 65 & 67 & 49.8 & 35.6 & 45 \\
alexrobertson4 & 18.8 & 18.8 & 5.6 & 51.4 & 7.8 & 7.8 & 12 & 12 & 5.2 & 5.2 & 14.4 & 12.2 & 60.8 & 69.2 & 12.4 & 12.4 \\
\end{tabular}
\end{adjustbox}
\end{subtable}

\vspace{5pt}

\begin{subtable}{\textwidth}
\centering
\setlength{\tabcolsep}{3pt}
\begin{adjustbox}{width=\textwidth}
\begin{tabular}{l*{16}{S}}
\toprule
\textbf{Team} &
\texttt{ja-JP} & \texttt{en-JP} & \texttt{ms-SG} & \texttt{zh-TW} &
\texttt{en-TW} & \texttt{es-EC} & \texttt{en-EC} & \texttt{sv-SE} &
\texttt{en-SE} & \texttt{tl-PH} & \texttt{en-PH} & \texttt{ta-SG} &
\texttt{ta-LK} & \texttt{en-LK} & \texttt{en-SG} & \texttt{Avg} \\
\midrule
\textbf{1. king001} & \bestmono68.2 & \bestmono56.2 & \bestmono79.4 & \bestmono62.2 & \bestmono54.8 & \bestmono69.8 & \bestmono58.6 & 62.8 & 56.8 & 58.4 & \bestmono63.6 & 51 & 36 & \bestmono52.2 & \bestmono82.2 & \bestmono59.5 \\
\rowcolor{gray!25}
\textbf{\texttt{GPT4.1}} & 66.4 & 54 & 76.6 & \bestllm63.4 & \bestllm57.4 & 53.2 & \bestllm61.2 & 61.6 & \bestllm59 & \bestllm64.4 & 61.2 & 65.2 & 38.6 & 47.4 & \bestllm85.2 & \bestllm59.5 \\
\textbf{2. K-NLPers }& 63.6 & 49.4 & 80.4 & 60.4 & \bestmono54.8 & 51.4 & 46.6 & 54.4 & 51.4 & \bestmono60 & 56.8 & \bestmono79.2 & \bestmono43.2 & 43.2 & 77.8 & 55.1 \\
\textbf{3. GUIR} & 63.6 & 48.8 & 72 & 61.2 & 51.4 & 63 & 57.4 & \bestmono63.4 & 53.4 & \bestmono60 & 53.2 & 42.8 & 29.4 & 47.4 & 79.8 & 54.4 \\
\rowcolor{gray!25}
\textbf{\texttt{Qwen3-2}} & 65 & 50.6 & 67 & 57 & 51.2 & 55.4 & 51.8 & 57.2 & 52.6 & 52 & 56 & 44.6 & 33.4 & 43 & 79 & 52 \\
\textbf{4. LocuPrompt} & 54.8 & 51.6 & 64.2 & 29.2 & 51.2 & 54.6 & 52.4 & 40 & 37 & 56.4 & 53.6 & 47.2 & 26.4 & 31.4 & 74 & 48.3 \\
\textbf{5. wangkongqiang} & 60.8 & 52.6 & 66.4 & 63 & 46.6 & 59.6 & 40.6 & 57.8 & 51.2 & 52.4 & 56.6 & 24.8 & 13.4 & 42 & 79.8 & 50.3 \\
\textbf{6. FLANS} & 28.2 & 35.4 & 26.4 & 0 & 32.4 & 22.6 & 31.4 & 0.2 & 36.8 & 23 & 40 & 0.2 & 0.2 & 30.2 & 61 & 24.5 \\
\textbf{7. Agentic} & 28.4 & {-} & 43.4 & 34.6 & {-} & 26.8 & {-} & 29.4 & {-} & 28.2 & {-} & 16.4 & 4 & {-} & 53.8 & 28.2 \\
qinchihongye & 66.6 & 54.6 & 75.2 & 64.8 & 54.8 & 66.8 & 57 & 62.8 & 57.8 & 57.6 & 59.4 & 53.2 & 37.4 & 48 & 79.6 & 58.3 \\
rustycoder & 64.4 & 47.8 & 58.4 & 66.8 & 50.2 & 55.6 & 48.8 & 57.8 & 51 & 55 & 50.2 & 33.6 & 31 & 41.2 & 77.8 & 51.1 \\
fedikallel & 61 & 43.8 & 65.2 & 60.4 & 49.2 & 53.4 & 41.4 & 55 & 42.4 & 52.2 & 48.2 & 57.6 & 32.4 & 39.8 & 72.4 & 48.7 \\
alexrobertson4 & 17.4 & 64.2 & 14.4 & 15.4 & 15.4 & 3.8 & 3.8 & 14.6 & 14.4 & {-} & {-} & 14.4 & 8.4 & 8.2 & 14.4 & 18.4 \\
\bottomrule
\end{tabular}
\end{adjustbox}
\end{subtable}

\caption{\textbf{SAQ evaluation scores across language–region pairs on the newly constructed \ours{} datasets, along with two baselines (\texttt{GPT-4.1} and \texttt{Qwen3-2}):}
Arabic (EG, MA, SA), English (EG, MA, SA, PV, BG, AU, FR, IE, JP, TW, EC, SE, PH, LK, SG), Basque (PV), Bulgarian (BG), Chinese (SG, TW), French (FR), Irish (IE), Japanese (JP), Malay (SG), Spanish (EC), Swedish (SE), Tagalog (PH), Tamil (SG, LK), and average (Avg). The best-performing results are highlighted in blue, and cells are highlighted in orange when an LLM baseline outperforms the submitted systems. Only teams with names in bold (i.e., those that submitted system description papers) are ranked; their citations can be found in the Appendix.}\label{tab:track_a}
\end{table*}

\begin{table*}[!ht]
\centering
\scriptsize

\begin{subtable}{\textwidth}
\centering
\begin{adjustbox}{width=\textwidth}
\begin{tabular}{l*{7}{S}}
\toprule
\textbf{Team} &
\texttt{ar-EG} & \texttt{ar-MA} & \texttt{ar-SA} & \texttt{bg-BG} &
\texttt{en-AU} & \texttt{es-EC} & \texttt{eu-PV} \\\midrule
\textbf{1. GUIR }& 91.03 & \bestmono91.53 & 87.16 & \bestmono99.54 & \bestmono94.54 & 98.67 & \bestmono94.23 \\
\textbf{2. uir-cis-7} & 96.74 & 83.98 & 91.89 & 99.54 & 90.06 & 98.36 & 97.12 \\
\textbf{3. Pinetree} & \bestmono94.57 & 81.95 & \bestmono93.02 & 98.77 & 93.18 & \bestmono99.28 & 91.63 \\
\rowcolor{gray!25}
\textbf{\texttt{GPT4.1}} & 94.29& 84.90 & 89.41 & 99.23 & 90.25 & 99.18 & 93.21 \\
\textbf{4. K-NLPers} & 91.03 & 80.85 & 81.31 & 95.68 & 92.59 & 98.57 & 90.79 \\
\rowcolor{gray!25}
\textbf{\texttt{Qwen3-2}} & 89.95 & 78.45 & 82.21 & 96.76 & 92.20 & 96.83 & 91.07 \\
\textbf{5. DFKI-MTL}& 94.84 & 79.93 & 80.41 & 94.60 & 90.84 & 97.54 & 87.16 \\
\textbf{6. wangkongqiang} & 93.21 & 76.43 & 80.86 & 92.90 & 84.41 & 96.52 & 87.72 \\
\textbf{7. Models without Borders }& 83.42 & 72.56 & 71.17 & 89.97 & 80.90 & 92.94 & 83.26 \\
\textbf{8. Agentic }& 85.05 & 75.69 & 68.92 & 88.58 & 80.70 & 90.07 & 80.65 \\
\textbf{9. CultRAG }& 80.43 & 76.98 & 74.32 & 87.81 & 82.65 & 86.28 & 78.98 \\
\textbf{10. LocuPrompt} & 73.91 & 67.59 & 74.77 & 75.31 & 86.94 & 92.94 & 76.37 \\
\textbf{11. Simorgh }& 68.75 & 63.35 & 61.04 & 81.94 & 84.99 & 90.28 & 75.53 \\
\textbf{12. FLANS} & {-} & {-} & {-} & {-} & 71.54 & 90.58 & {-} \\
 VerbanexAI & 7.60 & 2.20 & 6.00 & 10.00 & 31.60 & 5.20 & 1.00 \\
UTD-HLT & {-} & {-} & {-} & {-} & {-} & {-} & {-} \\
rustycoder & 93.21 & 87.66 & 90.09 & 99.69 & 92.01 & 98.26 & 92.74 
\\
fedikallel & 92.39 & 82.87 & 79.50 & 96.91 & 90.64 & 95.19 & 90.60 \\
narjes & {-} & {-} & {-} & {-} & 78.95 & {-} & {-} \\
\end{tabular}
\end{adjustbox}
\end{subtable}

\vspace{5pt}

\begin{subtable}{\textwidth}
\centering
\begin{adjustbox}{width=\textwidth}
\begin{tabular}{l*{8}{S}}
\toprule
\textbf{Team }&
\texttt{fr-FR} & \texttt{ga-IE} & \texttt{ja-JP} & \texttt{sv-SE} &
\texttt{ta-LK} & \texttt{tl-PH} & \texttt{zh-SG} & \texttt{Avg} \\
\midrule
\textbf{1. GUIR}& \bestmono98.70 & \bestmono98.48 & \bestmono91.71 & 94.41 & \bestmono97.13 & \bestmono96.53 & 93.69 & \bestmono94.81 \\
\textbf{2. uir-cis-7 }& 97.72 & 96.26 & 87.32 & \bestmono94.63 & 96.86 & 94.35 & \bestmono95.09 & 94.28 \\
\textbf{3. Pinetree }& 98.05 & 94.98 & 86.83 & 93.74 & 93.54 & 91.79 & 92.29 & 93.12 \\
\rowcolor{gray!25}
\textbf{\texttt{GPT4.1}} &97.39& 91.12 & 89.02 & 92.39 & 92.72 & 91.48 & 92.99 & 92.69 \\
\textbf{4. K-NLPers} & 96.09 & 91.24 & 85.85 & 87.47 & 95.96 & 88.47 & 91.82 & 90.55 \\
\rowcolor{gray!25}
\textbf{\texttt{Qwen3-2}} & 96.42 & 87.38 & 87.80 & 87.92 & 91.92 & 86.51 & 88.78 & 89.59 \\
\textbf{5. DFKI-MTL} & 93.81 & 90.89 & 88.78 & 82.55 & 94.61 & 84.10 & 88.32 & 89.17 \\
\textbf{6. wangkongqiang}& 93.16 & 87.62 & 84.15 & 89.04 & 91.38 & 80.48 & 87.62 & 87.54 \\
\textbf{7. Models without Borders}& 88.60 & 75.12 & 77.56 & 84.56 & 87.79 & 75.96 & 87.62 & 82.25 \\
\textbf{8. Agentic}& 86.64 & 84.58 & 80.24 & 78.08 & 84.74 & 72.04 & 84.11 & 81.44 \\
\textbf{9. CultRAG}& 86.64 & 87.50 & 80.98 & 79.87 & 89.77 & 77.77 & 88.08 & 82.72 \\
\textbf{10. LocuPrompt} & 86.32 & 80.72 & 84.15 & 71.36 & 81.69 & 70.16 & 81.07 & 78.81 \\
\textbf{11. Simorgh} & 78.83 & 85.63 & 78.05 & 68.23 & 84.56 & 77.84 & 79.21 & 77.02 \\
\textbf{12. FLANS }& {-} & {-} & {-} & {-} & {-} & {-} & 83.41 & 81.84 \\
VerbanexAI & 20.40 & 0.20 & 8.20 & 13.60 & 3.20 & 14.60 & 19.40 & 10.23 \\ 
UTD-HLT & {-} & {-} & 86.34 & {-} & {-} & {-} & {-} & 86.34 \\
rustycoder & 98.70 & 96.73 & 89.76 & 92.39 & 96.32 & 93.59 & 93.93 & 93.93 \\
fedikallel & 95.44 & 88.32 & 85.85 & 87.47 & 90.39 & 81.46 & 88.32 & 88.95 \\

narjes & {-} & {-} & 75.37 & {-} & {-} & {-} & {-} & 77.16 \\
\bottomrule
\end{tabular}
\end{adjustbox}
\end{subtable}

\caption{\textbf{MCQ evaluation scores across language–region pairs on the newly constructed \ours{} datasets along with two baselines (\texttt{GPT4.1} and \texttt{Qwen3-2}):}
Arabic (EG, MA, SA), English (AU), Basque (PV), Bulgarian (BG), Chinese (SG), French (FR), Irish (IE), Japanese (JP), Spanish (EC), Swedish (SE), Tagalog (PH), Tamil (LK), overall. The best-performing results are highlighted in blue, and cells are highlighted in orange when an LLM baseline outperforms the submitted systems. Only teams with names in bold (i.e., those that submitted system description papers) are ranked; their citations can be found in the Appendix.}
\label{tab:track_b}
\end{table*}

\paragraph{Track A: Short-Answer Questions (SAQ).}
Participants submit systems aimed at generating answers to short-answer questions. We assess their ability to generate accurate responses while accounting for cultural and linguistic diversity. This track includes two variants: (1) an English variant, in which answers are provided in English; and (2) a native-language variant, in which answers are provided in the original language.

\paragraph{Track B: Multiple-Choice Questions (MCQ).}
Participants submit systems aimed at answering questions provided in English for each target country or region. We evaluate the ability of the systems to select the culturally appropriate answer for each target region. In this track, each question includes four answer options, each representing a cultural perspective from a different country or region. The correct answer corresponds to the option that received the highest number of votes for the corresponding region during data collection.

\subsection{Evaluation Metrics}
\paragraph{Track A: SAQ}
We evaluate submitted responses by comparing them against gold-standard answers using exact matching, complemented by language-specific normalisation techniques. These include lemmatisation and stemming for highly inflectional languages (e.g., Arabic), as well as accent removal for languages such as Spanish, French, and Greek. This approach accounts for linguistic variation and ensures better alignment between human annotations and model outputs. Note that we intentionally do not use semantic similarity metrics, as, in culturally sensitive contexts, responses that are semantically similar may still be incorrect (e.g., ``having \textit{a croissant} for breakfast'' versus ``having \textit{cake/bread}''). Despite its limitations, exact matching provides a more reliable measure of correctness in this setting.

\paragraph{Track B: MCQ}
We evaluate whether submitted responses match the gold answers (i.e., the correct option for each question). In general, models are expected to achieve higher performance on MCQ than on SAQ, as selecting from predefined options is less challenging than generating free-form answers.

\section{Evaluation}
Our task attracted more than 140 registered participants, with more than 880 submissions in total. The official results in Tables \ref{tab:track_a} and \ref{tab:track_b} include the ranked final submissions of the teams that submitted a system description paper.

\subsection{Overview}
We note that the MCQ track (B) received the highest number of submissions, while the SAQ track (A) was the most challenging. Most teams submitted models for a wide range of languages, including underserved ones, which were comparable in terms of popularity.
We received final official submissions from 22 different teams. Performance varies considerably across languages. To ensure fairness, we report the best results on the newly constructed datasets in the main text, as some LLMs may have been fine-tuned on our \ours{} benchmark, which was publicly released in 2024.

\paragraph{Baselines}
To establish reference performance, we evaluate two LLM baselines: \texttt{GPT-4.1} \cite{openai2024gpt4} and \texttt{Qwen3-2} \cite{qwen}. Both models are prompted in zero-shot settings using simple task-specific instructions. For the short-answer question (SAQ) task, the models are instructed to produce a single concise answer without explanation. Similarly, for the multiple-choice question (MCQ) task, the models are asked to select one option from the provided alternatives (A--D) without additional explanation.

\subsection{Track A (SAQ) Results}

\subsubsection{Best-Performing Systems}

\paragraph{Team king001}
Team king001 \cite{king001_blend2026} implemented a Retrieval-Augmented Generation (RAG) architecture for cross-lingual cultural question answering. The system first parses each query to identify key elements, including the target region, question type, and core semantic intent. It then encodes both the query and knowledge-base documents using embeddings to retrieve the top-$k$ most relevant region-specific cultural knowledge fragments via vector similarity search. The retrieved passages are combined with the original query and provided to \texttt{GPT-5.2-chat}, which is prompted to generate concise responses that are both factually grounded and culturally and linguistically appropriate for the target region.

\paragraph{Team K-NLPers}
Team K-NLPers \cite{knlpers_blend2026} proposed a multi-agent framework based on a continent-level debate architecture that leverages culture-specific performance differences rather than relying on a single model. For the short-answer question answering task, their system uses three agents: a general-purpose model, a continent-specific model, and a country-level or culturally adjacent model. These agents generate answers independently, iteratively refine each other's outputs, and finally produce a decision through an adjudication stage that selects the most appropriate response.

\paragraph{Team LocuPrompt}
Team LocuPrompt \cite{locuprompt_blend2026} prompted locale-specific models and selected the best-performing LLM for each locale from a diverse pool, including \texttt{Llama~3}, \texttt{Qwen}, \texttt{DeepSeek}, \texttt{GPT-5}, and \texttt{Gemini}. Questions were first answered in English using culturally grounded prompts and subsequently back-translated into the target language. The team show that the Google Translate API helps improve performance for low-resource languages such as Irish and Basque. They further conducted a qualitative analysis on randomly sampled outputs and reported common model overgeneralisations (e.g., stereotypes related to Japanese culture).

\subsubsection{Takeaways}
In Track A (SAQ), we observe strong performance from our simple \texttt{GPT-4.1} baseline, which outperforms submitted systems in several languages, including Arabic variants (EG and SA), Basque, Bulgarian, Chinese (Singapore), French, Irish, Mandarin (Taiwan), Tagalog, and English (Australian and Singaporean). Interestingly, the top system achieves overall performance comparable to the \texttt{GPT-4.1} baseline, with only slight improvements.

Submitted systems typically generated short free-text responses using models such as \texttt{Gemini}, \texttt{GPT}, \texttt{Qwen}, \texttt{DeepSeek}, or \texttt{LLaMA}. Many approaches placed strong emphasis on prompt engineering by, for example, framing prompts from the perspective of a local resident. Systems also incorporated multilingual strategies, including pivot translation or direct prompting in the target language, as well as strict formatting controls, such as 1--3 word answer constraints, deterministic decoding, and grammar-checking post-processing.

Several submissions further adopted ensemble or multi-agent debate frameworks, in which general-purpose and region-specific models produced candidate answers that were aggregated through voting mechanisms or a judge model. Some teams also explored fine-tuning approaches, such as LoRA with synthetic training data, while a smaller number investigated more experimental techniques, such as activation steering using language vectors.

Only a few teams focused on a limited subset of languages (e.g., HausaNLP for Hausa; \citealp{hausanlp_blend2026}) or explicitly incorporated language-specific knowledge into their methodological design.

\subsection{Track B (MCQ) Results}

\subsubsection{Best-Performing Systems}

\paragraph{Team GUIR}
Team GUIR \cite{guir_blend2026} built a three-stage pipeline for the MCQ task: (1) zero-shot chain-of-thought inference using \texttt{GPT-5-mini}, (2) cross-locale majority voting to correct inconsistent predictions, and (3) a multi-agent debate protocol in which three LLM instances argue over and adjudicate remaining errors. This pipeline achieved 97.47\% overall accuracy across 30 locales, ranking first among all submitted systems in the MCQ track. The team also conducted a targeted human evaluation for Farsi (Iran) to highlight limitations of lemma-based evaluation for morphologically rich languages such as Persian. They reported that the lemma-matching scorer systematically underestimated model performance, with human annotators rating the system 18 percentage points higher.

\paragraph{Team K-NLPers} 
Team K-NLPers \cite{knlpers_blend2026} applied a continent-based multi-agent debate framework similar to their approach in Track A. However, due to the simpler structure of the MCQ task, their system relied primarily on high-performing general-purpose models and a streamlined debate protocol to reach the final decision.

\paragraph{Team uir-cis-7}
Team uir-cis-7 \cite{uircis7_blend2026} addressed the challenge by proposing a zero-shot Chain-of-Thought (CoT) framework that uses a reasoning-focused LLM configured with locale-conditioned personas to evaluate answer options before selecting the final response. Answer extraction was handled through a multi-stage regex-based strategy designed to ensure robust option formatting. The team also implemented a highly concurrent asynchronous pipeline to support large-scale inference. Their analysis highlights cultural blind spots in African and Middle Eastern contexts.

\paragraph{Team Pinetree}
Team Pinetree \cite{pinetree_blend2026} developed a multiple-choice reasoning system using models from the \texttt{Qwen} series together with \texttt{DeepSeek-V3.2-Exp}. The models generate the option they consider correct based on carefully designed prompts. The prompt incorporates lightweight behavioural constraints intended to mitigate common failure modes observed during development. In particular, the system encourages the model to prioritise widely practised everyday behaviours, favour nationally salient norms when regional variation exists, avoid selecting answers based purely on lexical overlap with the question, disregard exaggerated stereotypes unless they reflect genuine mainstream practices, and focus on contemporary everyday behaviour rather than historical or ceremonial customs. The team also provides an expanded error analysis and a taxonomy of cultural misalignments based on more than 5,000 cases.

\subsubsection{Takeaways}
Similar to the SAQ track, our \texttt{GPT-4.1} baseline performs strongly. However, it is generally slightly outperformed by the top teams (GUIR \cite{guir_blend2026}, uir-cis-7 \cite{uircis7_blend2026}, and Pinetree \cite{pinetree_blend2026}).

We note that while most teams experimented with multiple LLMs and languages, some submissions focused on specific models or languages  (e.g., FLANS; \citealp{flans_blend2026}). Furthermore, most teams relied on standard prompting strategies combined with different large language models. Systems typically used instruction prompting, light reasoning prompts (e.g., Chain-of-Thought), and structured answer extraction to ensure valid A/B/C/D outputs.

Across submissions, prompting and Retrieval-Augmented Generation (RAG) approaches dominated system design, with region-aware prompting emerging as the most common strategy, e.g., teams LocuPrompt \cite{locuprompt_blend2026} and Howard University \cite{howard_uni_blend2026}. Other teams explored complementary approaches, including persona-based prompting (e.g., teams UTD-HLTRI \cite{utd_hltri_blend2026} and uir-cis-7 \cite{uircis7_blend2026}), intent-conditioned prompting (e.g., team CultRAG \cite{culturag_blend2026}), and data augmentation techniques (e.g., teams DFKI-MLT \cite{dfki_mlt_blend2026} and Agentic \cite{agentic_blend2026}).

\section{Discussion}
Beyond system performance, several teams conducted additional analyses that provide valuable insights into model limitations, cultural biases, and evaluation challenges, which we summarise below.

\paragraph{Evaluation Challenges}
We observed marked differences in SAQ performance scores across language--culture pairs when comparing scores achieved on the earlier version of our \ours{} benchmark with those on the newly extended datasets. Given the high performance reported by some teams, we considered the possibility of data contamination; it is plausible that \texttt{Gemini} and \texttt{GPT-5} had been fine-tuned on earlier versions of our \ours{} benchmark. This is particularly evident as some submissions achieved perfect scores in certain locales (e.g., zh-CN by teams king001 and LocuPrompt) and near-perfect scores in others, such as am-ET and ar-DZ. In the main text, we report the best results on the newly constructed datasets, acknowledging that some LLMs may have been fine-tuned on our \ours{} benchmark, which was publicly released in 2024.

Assessing SAQs remains challenging, particularly for non-English languages. For instance, team GIUR \cite{guir_blend2026}, who conducted a targeted human evaluation for Farsi (Iran), highlighted the limitations of lemma-based evaluation for morphologically rich languages such as Persian. They found that the lemma-matching scorer systematically underestimated model performance, with human annotators rating the system 18 percentage points higher.While we have made every effort to assess performance fairly, evaluating short answers remains an open problem, particularly when semantic similarity cannot be relied upon due to the risk of semantically similar yet incorrect responses relative to the gold answers.

\paragraph{Challenges for Underrepresented Languages and Cultures}
Some submissions highlighted the difficulties current models face when dealing with underrepresented languages and cultural contexts. For example, HausaNLP focused exclusively on Hausa \cite{hausanlp_blend2026}. Their results demonstrate a substantial performance gap when prompting models directly in low-resource languages such as Hausa, consistent with previous findings \cite{myung2024blend}. Similarly, teams Simorgh \cite{simorgh_blend2026} and uir-cis-7 \cite{uircis7_blend2026} focused on limitations in non-Western cultural contexts, examining, for instance, regional variation in Arabic and Japanese cultural knowledge.

\paragraph{Cultural Misalignment and Model Behaviour}
Several teams investigated sources of cultural misalignment and behavioural patterns in model outputs. For instance, team Models without Borders \cite{models_without_borders_blend2026} explored geographically and linguistically conditioned web retrieval as a lightweight alternative to training-intensive cultural adaptation, while team FlANS \cite{flans_blend2026} examined the performance of smaller models for more sustainable and efficient NLP systems. Other submissions provided detailed error analyses. For example, team GUIR \cite{guir_blend2026} analysed the limitations of evaluation pipelines relying on lemmatisers for morphologically rich or low-resource languages, and team Pinetree \cite{pinetree_blend2026} examined thousands of cases of cultural misalignment. 

Additional observations include cultural overgeneralisation reported by team LocuPrompt \cite{locuprompt_blend2026} and sensitivity to option ordering in MCQ settings noted by team uir-cis-7 \cite{uircis7_blend2026}. This latter finding reflects a key property of the dataset: identical English questions frequently appear across multiple regional variants with different answer options, requiring models to perform region-specific reasoning rather than relying on memorised answers.

\paragraph{Methodological Insights on Modelling Choices}
Other teams carried out ablation studies to better understand system behaviour. Team CultRAG \cite{culturag_blend2026} showed that RAG improves performance for underrepresented cultures (e.g., Japan and Ethiopia) but can sometimes reduce performance in high-resource Western contexts (e.g., the UK and US), highlighting the importance of retrieval coverage and answer distribution biases. They also proposed several directions for future work, including confidence-conditioned selective retrieval, targeted knowledge-base expansion, intent-aware retrieval, and cross-encoder reranking.

Similarly, the uir-cis-7 team \cite{uircis7_blend2026} evaluated the limitations of persona-based prompting through controlled experiments.

Finally, DFKI-MLT \cite{dfki_mlt_blend2026} investigated activation steering for cultural reasoning tasks. Their analysis shows that steering gains are highly layer-sensitive and vary across language–region pairs, while prompt design interacts strongly with steering effectiveness, suggesting that prompt design and activation steering should be treated as a jointly optimised inference-time adaptation strategy rather than independent components.

\section{Conclusion}

We presented our shared task on evaluating the adaptability of large language models (LLMs) and NLP systems across multiple languages and cultures, covering more than 30 language–culture pairs, predominantly representing low-resource languages spoken across multiple continents. Submitted systems were ranked based on the match between predicted labels and gold labels in two tracks: (a)\ short-answer questions (SAQ) and (b)\ multiple-choice questions (MCQ).

We summarised the reported results, discussing the predominant and best-performing methods, as well as shared analyses and insights into model behaviour, cultural misalignment, and modelling choices. Overall, performance varied substantially between the two tracks, with MCQ being considerably less challenging than SAQ, and across locales (e.g., simple prompting can be sufficient for Singaporean English short-answer questions but not for Irish).

We highlight several open challenges, particularly for underrepresented cultures and languages, as well as for low-resource settings.

\section*{Limitations} \label{sec:limitations}

We do not claim that our benchmark is fully representative of all language--culture pairs. We also acknowledge the inherent complexity and nuance involved in defining culture \cite{zhou2025culture}, as well as the fact that our use of the term within national borders is narrower than broader anthropological or sociolinguistic definitions \cite{alkhamissi-etal-2026-hire}. However, for reasons of annotation feasibility and scalability, we adopt practical design choices by restricting culture to these boundaries and limiting the benchmark to predefined topics within specific countries. We encourage the community to extend this work by covering additional regions and increasing the number of annotators per locale.

Additionally, evaluating short-answer questions presents notable challenges, as it is difficult to account for all valid lexical and morphological variations---particularly in highly inflectional languages such as Arabic and Persian, as also noted by team GUIR \cite{guir_blend2026}. We therefore encourage the exploration of alternative evaluation metrics that better capture variation in short-form answers.

Despite these limitations, we believe the dataset provides a useful starting point for research on cultural and linguistic adaptability in NLP systems.

\section*{Ethical Considerations}

Language leads involved in benchmark creation were academics proficient in English, which served as the reference language for communication and translation during the annotation process. While each language lead was a native speaker of the language they were responsible for and recruited annotators who had spent at least half of their lives in the relevant region, we do not claim that all cultural variation within each locale is fully captured. We also acknowledge that annotators may bring their own subtle, internalised perceptions and biases to the annotation process. Finally, we follow the recommendations of \citet{ousidhoum-etal-2025-building} for constructing datasets in low-resource languages. Specifically, all annotators involved in the study are native speakers and were compensated at rates exceeding the local minimum wage.

\section*{Acknowledgments}

Junho Myung, Jiho Jin, and Alice Oh were supported by LG AI Research and the Institute of Information \& Communications Technology Planning \& Evaluation (IITP) grant funded by the Korea government (MSIT) (No. RS-2024-00509258 and No. RS-2024-00469482, Global AI Frontier Lab).
\newline
Jose Camacho-Collados and part of this project were supported by a UKRI Future Leaders Fellowship.
\newline
Joseba Fernandez de Landa and Rodrigo Agerri were supported by Grant DeepThought (PID2024-159202OB-C21), funded by MICIU/AEI /10.13039/501100011033 and ERDF, EU, as well as the ALIA Models Development Project, Resolution of the SEDIA 19/08/2024, within the framework of the National Language Technologies Plan – ENIA 2024, funded by MTDFP, PRTR, and the EU – NextGenerationEU.
\newline
Joseph Marvin Imperial is supported by the National University of the Philippines and the UKRI Centre for Doctoral Training in Accountable, Responsible, and Transparent AI [EP/S023437/1] at the University of Bath.



\newpage

\bibliography{anthology,custom,semeval-task-7}

\begin{thebibliography}{37}
\providecommand{\natexlab}[1]{#1}

\bibitem[{Adam et~al.(2026)Adam, Aliyu, Aji, Abubakar, and Shuaibu}]{hausanlp_blend2026}
Faisal~Muhammad Adam, Lukman~Jibril Aliyu, Sani Aji, Abdulhamid Abubakar, and Aliyu~Rabiu Shuaibu. 2026.
\newblock {H}ausa{N}{L}{P} at {S}em{E}val-2026 {T}ask 7: {P}rompt-based {H}ausa {C}ultural {Q}uestion {A}nswering.
\newblock In \emph{Proceedings of the 20th International Workshop on Semantic Evaluation ({S}em{E}val-2026)}.

\bibitem[{Adilazuarda et~al.(2024)Adilazuarda, Mukherjee, Lavania, Singh, Aji, O{'}Neill, Modi, and Choudhury}]{adilazuarda-etal-2024-towards}
Muhammad~Farid Adilazuarda, Sagnik Mukherjee, Pradhyumna Lavania, Siddhant~Shivdutt Singh, Alham~Fikri Aji, Jacki O{'}Neill, Ashutosh Modi, and Monojit Choudhury. 2024.
\newblock \href {https://doi.org/10.18653/v1/2024.emnlp-main.882} {Towards measuring and modeling {\textquotedblleft}culture{\textquotedblright} in {LLM}s: A survey}.
\newblock In \emph{Proceedings of the 2024 Conference on Empirical Methods in Natural Language Processing}, pages 15763--15784, Miami, Florida, USA. Association for Computational Linguistics.

\bibitem[{Adjei and Aryal(2026)}]{howard_uni_blend2026}
Isaac~Nyadu Adjei and Saurav~K. Aryal. 2026.
\newblock {H}oward {U}niversity-{A}{I}4{P}{C} at {S}em{E}val-2026 {T}ask 7: {C}ulturally {A}ware {M}ultilingual {M}odel {R}outing {T}hrough a {M}ixture-of-{S}pecialists {F}ramework.
\newblock In \emph{Proceedings of the 20th International Workshop on Semantic Evaluation ({S}em{E}val-2026)}.

\bibitem[{Al~Ghussin et~al.(2026)Al~Ghussin, Gurgurov, Hamidullah, van Genabith, España-Bonet, and Ostermann}]{dfki_mlt_blend2026}
Yusser Al~Ghussin, Daniil Gurgurov, Yasser Hamidullah, Josef van Genabith, Cristina España-Bonet, and Simon Ostermann. 2026.
\newblock {D}{F}{K}{I}-{M}{L}{T} at {S}em{E}val-2026 {T}ask 7: {S}teering {M}ultilingual {M}odels {T}owards {C}ultural {K}nowledge.
\newblock In \emph{Proceedings of the 20th International Workshop on Semantic Evaluation ({S}em{E}val-2026)}.

\bibitem[{Alkhamissi et~al.(2026)Alkhamissi, Xiao, AlKhamissi, and Diab}]{alkhamissi-etal-2026-hire}
Mai Alkhamissi, Yunze Xiao, Badr AlKhamissi, and Mona~T. Diab. 2026.
\newblock Hire your anthropologist! rethinking culture benchmarks through an anthropological lens.
\newblock In \emph{Findings of the {A}ssociation for {C}omputational {L}inguistics: {EACL} 2026}, pages 1218--1235. Association for Computational Linguistics.

\bibitem[{Almanza et~al.(2026)Almanza, Serrano, Puertas, and Martinez~Santos}]{verbanexAI_blend2026}
Danileth Almanza, Jairo Serrano, Edwin Puertas, and Juan~Carlos Martinez~Santos. 2026.
\newblock {V}erba{N}ex{A}{I} at {S}em{E}val-2026 {T}ask 7: {I}ntegrating {W}eb {S}nippets and {R}{A}{G} for the {E}valuation of {M}ultilingual {C}ultural {K}nowledge in {L}{L}{M}s.
\newblock In \emph{Proceedings of the 20th International Workshop on Semantic Evaluation ({S}em{E}val-2026)}.

\bibitem[{Bai et~al.(2023)Bai, Bai, Chu, Cui, Dang, Deng, Fan, Ge, Han, Huang, Hui, Ji, Li, Lin, Lin, Liu, Liu, Lu, Lu, Ma, Men, Ren, Ren, Tan, Tan, Tu, Wang, Wang, Wang, Wu, Xu, Xu, Yang, Yang, Yang, Yang, Yao, Yu, Yuan, Yuan, Zhang, Zhang, Zhang, Zhang, Zhou, Zhou, Zhou, and Zhu}]{qwen}
Jinze Bai, Shuai Bai, Yunfei Chu, Zeyu Cui, Kai Dang, Xiaodong Deng, Yang Fan, Wenbin Ge, Yu~Han, Fei Huang, Binyuan Hui, Luo Ji, Mei Li, Junyang Lin, Runji Lin, Dayiheng Liu, Gao Liu, Chengqiang Lu, Keming Lu, Jianxin Ma, Rui Men, Xingzhang Ren, Xuancheng Ren, Chuanqi Tan, Sinan Tan, Jianhong Tu, Peng Wang, Shijie Wang, Wei Wang, Shengguang Wu, Benfeng Xu, Jin Xu, An~Yang, Hao Yang, Jian Yang, Shusheng Yang, Yang Yao, Bowen Yu, Hongyi Yuan, Zheng Yuan, Jianwei Zhang, Xingxuan Zhang, Yichang Zhang, Zhenru Zhang, Chang Zhou, Jingren Zhou, Xiaohuan Zhou, and Tianhang Zhu. 2023.
\newblock \href {https://arxiv.org/abs/2309.16609} {Qwen technical report}.
\newblock \emph{arXiv preprint arXiv:2309.16609}.

\bibitem[{Bogdanova et~al.(2026)Bogdanova, Sun, Han, Amat-Lefort, and Plaza-del Arco}]{flans_blend2026}
Liliia Bogdanova, Shiran Sun, Lifeng Han, Natalia Amat-Lefort, and Flor~Miriam Plaza-del Arco. 2026.
\newblock {F}{L}{A}{N}{S} at {S}em{E}val-2026 {T}ask 7: {R}{A}{G} with {O}pen-{S}ourced {S}maller {L}{L}{M}s for {E}veryday {K}nowledge {A}cross {D}iverse {L}anguages and {C}ultures.
\newblock In \emph{Proceedings of the 20th International Workshop on Semantic Evaluation ({S}em{E}val-2026)}.

\bibitem[{Durmus et~al.(2023)Durmus, Nyugen, Liao, Schiefer, Askell, Bakhtin, Chen, Hatfield-Dodds, Hernandez, Joseph et~al.}]{durmus2023towards}
Esin Durmus, Karina Nyugen, Thomas~I Liao, Nicholas Schiefer, Amanda Askell, Anton Bakhtin, Carol Chen, Zac Hatfield-Dodds, Danny Hernandez, Nicholas Joseph, et~al. 2023.
\newblock \href {https://arxiv.org/abs/2306.16388} {Towards measuring the representation of subjective global opinions in language models}.
\newblock \emph{arXiv preprint arXiv:2306.16388}.

\bibitem[{Fung et~al.(2024)Fung, Zhao, Doo, Sun, and Ji}]{fung2024massively}
Yi~Fung, Ruining Zhao, Jae Doo, Chenkai Sun, and Heng Ji. 2024.
\newblock \href {https://arxiv.org/abs/2402.09369} {Massively multi-cultural knowledge acquisition \& lm benchmarking}.
\newblock \emph{arXiv preprint arXiv:2402.09369}.

\bibitem[{Gao et~al.(2026)Gao, Mao, Shi, Zhaxi, Sun, Li, and Li}]{uircis7_blend2026}
Jianning Gao, Xianling Mao, Shumin Shi, Duanzhi Zhaxi, Yingbo Sun, Xiandeng Li, and Binyang Li. 2026.
\newblock uir-cis-7 at {S}em{E}val-2026 {T}ask 7: {Z}ero-{S}hot {C}hain-of-{T}hought {R}easoning for {C}ross-{C}ultural {D}aily {K}nowledge.
\newblock In \emph{Proceedings of the 20th International Workshop on Semantic Evaluation ({S}em{E}val-2026)}.

\bibitem[{Iranmanesh et~al.(2026)Iranmanesh, Frieder, and Goharian}]{guir_blend2026}
Reihaneh Iranmanesh, Ophir Frieder, and Nazli Goharian. 2026.
\newblock {G}{U}{I}{R} at {S}em{E}val-2026 {T}ask 7: {P}robing {C}ultural {K}nowledge in {L}{L}{M}s via {M}ulti-{A}gent {D}ebate.
\newblock In \emph{Proceedings of the 20th International Workshop on Semantic Evaluation ({S}em{E}val-2026)}.

\bibitem[{Jin et~al.(2026)Jin, Meng, Yin, Jiang, and Li}]{king001_blend2026}
Meizhi Jin, Zhichao Meng, Junqi Yin, Lianxin Jiang, and Jianyu Li. 2026.
\newblock king001 at {S}em{E}val-2026 {T}ask 7: {C}ross-{L}anguage {C}ultural {E}veryday {K}nowledge {Q}\&{A} {S}ystem {B}ased on {R}{A}{G}.
\newblock In \emph{Proceedings of the 20th International Workshop on Semantic Evaluation ({S}em{E}val-2026)}.

\bibitem[{Kaneko et~al.(2024)Kaneko, Bollegala, Okazaki, and Baldwin}]{kaneko2024evaluating}
Masahiro Kaneko, Danushka Bollegala, Naoaki Okazaki, and Timothy Baldwin. 2024.
\newblock \href {https://arxiv.org/abs/2401.15585} {Evaluating gender bias in large language models via chain-of-thought prompting}.
\newblock \emph{arXiv preprint arXiv:2401.15585}.

\bibitem[{Kim et~al.(2024)Kim, Suk, Oh, Yoo, Thorne, and Oh}]{kim2024click}
Eunsu Kim, Juyoung Suk, Philhoon Oh, Haneul Yoo, James Thorne, and Alice Oh. 2024.
\newblock Click: A benchmark dataset of cultural and linguistic intelligence in korean.
\newblock In \emph{Proceedings of the 2024 Joint International Conference on Computational Linguistics, Language Resources and Evaluation (LREC-COLING 2024)}, pages 3335--3346.

\bibitem[{Koto et~al.(2024)Koto, Mahendra, Aisyah, and Baldwin}]{koto2024indoculture}
Fajri Koto, Rahmad Mahendra, Nurul Aisyah, and Timothy Baldwin. 2024.
\newblock \href {https://arxiv.org/abs/2404.01854} {Indoculture: Exploring geographically-influenced cultural commonsense reasoning across eleven indonesian provinces}.
\newblock \emph{arXiv preprint arXiv:2404.01854}.

\bibitem[{Liu et~al.(2025)Liu, Gurevych, and Korhonen}]{liu2025culturallyawareadaptednlp}
Chen~Cecilia Liu, Iryna Gurevych, and Anna Korhonen. 2025.
\newblock \href {https://arxiv.org/abs/2406.03930} {Culturally aware and adapted nlp: A taxonomy and a survey of the state of the art}.
\newblock \emph{Preprint}, arXiv:2406.03930.

\bibitem[{Myung et~al.(2024)Myung, Lee, Zhou, Jin, Putri, Antypas, Borkakoty, Kim, Perez-Almendros, Ayele, Guti\'{e}rrez-Basulto, Ib\'{a}\~{n}ez Garc\'{\i}a, Lee, Muhammad, Park, Rzayev, White, Yimam, Pilehvar, Ousidhoum, Camacho-Collados, and Oh}]{myung2024blend}
Junho Myung, Nayeon Lee, Yi~Zhou, Jiho Jin, Rifki~Afina Putri, Dimosthenis Antypas, Hsuvas Borkakoty, Eunsu Kim, Carla Perez-Almendros, Abinew~Ali Ayele, V\'{\i}ctor Guti\'{e}rrez-Basulto, Yazm\'{\i}n Ib\'{a}\~{n}ez Garc\'{\i}a, Hwaran Lee, Shamsuddeen~Hassan Muhammad, Kiwoong Park, Anar~Sabuhi Rzayev, Nina White, Seid~Muhie Yimam, Mohammad~Taher Pilehvar, Nedjma Ousidhoum, Jose Camacho-Collados, and Alice Oh. 2024.
\newblock \href {https://proceedings.neurips.cc/paper_files/paper/2024/file/8eb88844dafefa92a26aaec9f3acad93-Paper-Datasets_and_Benchmarks_Track.pdf} {Blend: A benchmark for llms on everyday knowledge in diverse cultures and languages}.
\newblock In \emph{Advances in Neural Information Processing Systems}, volume~37, pages 78104--78146. Curran Associates, Inc.

\bibitem[{Naous et~al.(2024)Naous, Ryan, Ritter, and Xu}]{naous-etal-2024-beer}
Tarek Naous, Michael~J Ryan, Alan Ritter, and Wei Xu. 2024.
\newblock \href {https://doi.org/10.18653/v1/2024.acl-long.862} {Having beer after prayer? measuring cultural bias in large language models}.
\newblock In \emph{Proceedings of the 62nd Annual Meeting of the Association for Computational Linguistics (Volume 1: Long Papers)}, pages 16366--16393, Bangkok, Thailand. Association for Computational Linguistics.

\bibitem[{Navigli et~al.(2023)Navigli, Conia, and Ross}]{navigli2023biases}
Roberto Navigli, Simone Conia, and Bj{\"o}rn Ross. 2023.
\newblock Biases in large language models: origins, inventory, and discussion.
\newblock \emph{ACM Journal of Data and Information Quality}, 15(2):1--21.

\bibitem[{Ning(2026)}]{locuprompt_blend2026}
Jingke Ning. 2026.
\newblock {L}ocu{P}rompt at {S}em{E}val-2026 {T}ask 7: {A} {M}ultilingual {P}rompting {F}ramework for {C}ross-{C}ultural {E}veryday {K}nowledge in {L}{L}{M}s.
\newblock In \emph{Proceedings of the 20th International Workshop on Semantic Evaluation ({S}em{E}val-2026)}.

\bibitem[{OpenAI et~al.(2024)OpenAI, Achiam, Adler, Agarwal, Ahmad, Akkaya, Aleman, Almeida, Altenschmidt, Altman, Anadkat, Avila, Babuschkin, Balaji, Balcom, Baltescu, Bao, Bavarian, Belgum, Bello, Berdine, Bernadett-Shapiro, Berner, Bogdonoff, Boiko, Boyd, Brakman, Brockman, Brooks, Brundage, Button, Cai, Campbell, Cann, Carey, Carlson, Carmichael, Chan, Chang, Chantzis, Chen, Chen, Chen, Chen, Chen, Chess, Cho, Chu, Chung, Cummings, Currier, Dai, Decareaux, Degry, Deutsch, Deville, Dhar, Dohan, Dowling, Dunning, Ecoffet, Eleti, Eloundou, Farhi, Fedus, Felix, Fishman, Forte, Fulford, Gao, Georges, Gibson, Goel, Gogineni, Goh, Gontijo-Lopes, Gordon, Grafstein, Gray, Greene, Gross, Gu, Guo, Hallacy, Han, Harris, He, Heaton, Heidecke, Hesse, Hickey, Hickey, Hoeschele, Houghton, Hsu, Hu, Hu, Huizinga, Jain, Jain, Jang, Jiang, Jiang, Jin, Jin, Jomoto, Jonn, Jun, Kaftan, Łukasz Kaiser, Kamali, Kanitscheider, Keskar, Khan, Kilpatrick, Kim, Kim, Kim, Kirchner, Kiros, Knight, Kokotajlo, Łukasz Kondraciuk,
  Kondrich, Konstantinidis, Kosic, Krueger, Kuo, Lampe, Lan, Lee, Leike, Leung, Levy, Li, Lim, Lin, Lin, Litwin, Lopez, Lowe, Lue, Makanju, Malfacini, Manning, Markov, Markovski, Martin, Mayer, Mayne, McGrew, McKinney, McLeavey, McMillan, McNeil, Medina, Mehta, Menick, Metz, Mishchenko, Mishkin, Monaco, Morikawa, Mossing, Mu, Murati, Murk, Mély, Nair, Nakano, Nayak, Neelakantan, Ngo, Noh, Ouyang, O'Keefe, Pachocki, Paino, Palermo, Pantuliano, Parascandolo, Parish, Parparita, Passos, Pavlov, Peng, Perelman, de~Avila Belbute~Peres, Petrov, de~Oliveira~Pinto, Michael, Pokorny, Pokrass, Pong, Powell, Power, Power, Proehl, Puri, Radford, Rae, Ramesh, Raymond, Real, Rimbach, Ross, Rotsted, Roussez, Ryder, Saltarelli, Sanders, Santurkar, Sastry, Schmidt, Schnurr, Schulman, Selsam, Sheppard, Sherbakov, Shieh, Shoker, Shyam, Sidor, Sigler, Simens, Sitkin, Slama, Sohl, Sokolowsky, Song, Staudacher, Such, Summers, Sutskever, Tang, Tezak, Thompson, Tillet, Tootoonchian, Tseng, Tuggle, Turley, Tworek, Uribe, Vallone,
  Vijayvergiya, Voss, Wainwright, Wang, Wang, Wang, Ward, Wei, Weinmann, Welihinda, Welinder, Weng, Weng, Wiethoff, Willner, Winter, Wolrich, Wong, Workman, Wu, Wu, Wu, Xiao, Xu, Yoo, Yu, Yuan, Zaremba, Zellers, Zhang, Zhang, Zhao, Zheng, Zhuang, Zhuk, and Zoph}]{openai2024gpt4}
OpenAI, Josh Achiam, Steven Adler, Sandhini Agarwal, Lama Ahmad, Ilge Akkaya, Florencia~Leoni Aleman, Diogo Almeida, Janko Altenschmidt, Sam Altman, Shyamal Anadkat, Red Avila, Igor Babuschkin, Suchir Balaji, Valerie Balcom, Paul Baltescu, Haiming Bao, Mohammad Bavarian, Jeff Belgum, Irwan Bello, Jake Berdine, Gabriel Bernadett-Shapiro, Christopher Berner, Lenny Bogdonoff, Oleg Boiko, Madelaine Boyd, Anna-Luisa Brakman, Greg Brockman, Tim Brooks, Miles Brundage, Kevin Button, Trevor Cai, Rosie Campbell, Andrew Cann, Brittany Carey, Chelsea Carlson, Rory Carmichael, Brooke Chan, Che Chang, Fotis Chantzis, Derek Chen, Sully Chen, Ruby Chen, Jason Chen, Mark Chen, Ben Chess, Chester Cho, Casey Chu, Hyung~Won Chung, Dave Cummings, Jeremiah Currier, Yunxing Dai, Cory Decareaux, Thomas Degry, Noah Deutsch, Damien Deville, Arka Dhar, David Dohan, Steve Dowling, Sheila Dunning, Adrien Ecoffet, Atty Eleti, Tyna Eloundou, David Farhi, Liam Fedus, Niko Felix, Simón~Posada Fishman, Juston Forte, Isabella Fulford, Leo
  Gao, Elie Georges, Christian Gibson, Vik Goel, Tarun Gogineni, Gabriel Goh, Rapha Gontijo-Lopes, Jonathan Gordon, Morgan Grafstein, Scott Gray, Ryan Greene, Joshua Gross, Shixiang~Shane Gu, Yufei Guo, Chris Hallacy, Jesse Han, Jeff Harris, Yuchen He, Mike Heaton, Johannes Heidecke, Chris Hesse, Alan Hickey, Wade Hickey, Peter Hoeschele, Brandon Houghton, Kenny Hsu, Shengli Hu, Xin Hu, Joost Huizinga, Shantanu Jain, Shawn Jain, Joanne Jang, Angela Jiang, Roger Jiang, Haozhun Jin, Denny Jin, Shino Jomoto, Billie Jonn, Heewoo Jun, Tomer Kaftan, Łukasz Kaiser, Ali Kamali, Ingmar Kanitscheider, Nitish~Shirish Keskar, Tabarak Khan, Logan Kilpatrick, Jong~Wook Kim, Christina Kim, Yongjik Kim, Jan~Hendrik Kirchner, Jamie Kiros, Matt Knight, Daniel Kokotajlo, Łukasz Kondraciuk, Andrew Kondrich, Aris Konstantinidis, Kyle Kosic, Gretchen Krueger, Vishal Kuo, Michael Lampe, Ikai Lan, Teddy Lee, Jan Leike, Jade Leung, Daniel Levy, Chak~Ming Li, Rachel Lim, Molly Lin, Stephanie Lin, Mateusz Litwin, Theresa Lopez, Ryan
  Lowe, Patricia Lue, Anna Makanju, Kim Malfacini, Sam Manning, Todor Markov, Yaniv Markovski, Bianca Martin, Katie Mayer, Andrew Mayne, Bob McGrew, Scott~Mayer McKinney, Christine McLeavey, Paul McMillan, Jake McNeil, David Medina, Aalok Mehta, Jacob Menick, Luke Metz, Andrey Mishchenko, Pamela Mishkin, Vinnie Monaco, Evan Morikawa, Daniel Mossing, Tong Mu, Mira Murati, Oleg Murk, David Mély, Ashvin Nair, Reiichiro Nakano, Rajeev Nayak, Arvind Neelakantan, Richard Ngo, Hyeonwoo Noh, Long Ouyang, Cullen O'Keefe, Jakub Pachocki, Alex Paino, Joe Palermo, Ashley Pantuliano, Giambattista Parascandolo, Joel Parish, Emy Parparita, Alex Passos, Mikhail Pavlov, Andrew Peng, Adam Perelman, Filipe de~Avila Belbute~Peres, Michael Petrov, Henrique~Ponde de~Oliveira~Pinto, Michael, Pokorny, Michelle Pokrass, Vitchyr~H. Pong, Tolly Powell, Alethea Power, Boris Power, Elizabeth Proehl, Raul Puri, Alec Radford, Jack Rae, Aditya Ramesh, Cameron Raymond, Francis Real, Kendra Rimbach, Carl Ross, Bob Rotsted, Henri Roussez,
  Nick Ryder, Mario Saltarelli, Ted Sanders, Shibani Santurkar, Girish Sastry, Heather Schmidt, David Schnurr, John Schulman, Daniel Selsam, Kyla Sheppard, Toki Sherbakov, Jessica Shieh, Sarah Shoker, Pranav Shyam, Szymon Sidor, Eric Sigler, Maddie Simens, Jordan Sitkin, Katarina Slama, Ian Sohl, Benjamin Sokolowsky, Yang Song, Natalie Staudacher, Felipe~Petroski Such, Natalie Summers, Ilya Sutskever, Jie Tang, Nikolas Tezak, Madeleine~B. Thompson, Phil Tillet, Amin Tootoonchian, Elizabeth Tseng, Preston Tuggle, Nick Turley, Jerry Tworek, Juan Felipe~Cerón Uribe, Andrea Vallone, Arun Vijayvergiya, Chelsea Voss, Carroll Wainwright, Justin~Jay Wang, Alvin Wang, Ben Wang, Jonathan Ward, Jason Wei, CJ~Weinmann, Akila Welihinda, Peter Welinder, Jiayi Weng, Lilian Weng, Matt Wiethoff, Dave Willner, Clemens Winter, Samuel Wolrich, Hannah Wong, Lauren Workman, Sherwin Wu, Jeff Wu, Michael Wu, Kai Xiao, Tao Xu, Sarah Yoo, Kevin Yu, Qiming Yuan, Wojciech Zaremba, Rowan Zellers, Chong Zhang, Marvin Zhang, Shengjia
  Zhao, Tianhao Zheng, Juntang Zhuang, William Zhuk, and Barret Zoph. 2024.
\newblock \href {https://arxiv.org/abs/2303.08774} {Gpt-4 technical report}.
\newblock \emph{Preprint}, arXiv:2303.08774.

\bibitem[{Ousidhoum et~al.(2025)Ousidhoum, Beloucif, and Mohammad}]{ousidhoum-etal-2025-building}
Nedjma Ousidhoum, Meriem Beloucif, and Saif~M. Mohammad. 2025.
\newblock \href {https://doi.org/10.18653/v1/2025.acl-long.435} {Building better: Avoiding pitfalls in developing language resources when data is scarce}.
\newblock In \emph{Proceedings of the 63rd Annual Meeting of the Association for Computational Linguistics (Volume 1: Long Papers)}, pages 8881--8894, Vienna, Austria. Association for Computational Linguistics.

\bibitem[{Pawar et~al.(2024)Pawar, Park, Jin, Arora, Myung, Yadav, Haznitrama, Song, Oh, and Augenstein}]{pawar2024surveyculturalawarenesslanguage}
Siddhesh Pawar, Junyeong Park, Jiho Jin, Arnav Arora, Junho Myung, Srishti Yadav, Faiz~Ghifari Haznitrama, Inhwa Song, Alice Oh, and Isabelle Augenstein. 2024.
\newblock \href {https://arxiv.org/abs/2411.00860} {Survey of cultural awareness in language models: Text and beyond}.
\newblock \emph{Preprint}, arXiv:2411.00860.

\bibitem[{Rahman et~al.(2026)Rahman, Ailneni, and Harabagiu}]{utd_hltri_blend2026}
Mohammad~Marufur Rahman, Rakshitha~Rao Ailneni, and Sanda Harabagiu. 2026.
\newblock {U}{T}{D}-{H}{L}{T}{R}{I} at {S}em{E}val-2026 {T}ask 7: {B}ridging {C}ultural {K}nowledge {G}aps in {L}{L}{M}s via {W}eb-{A}ugmented {C}ontext.
\newblock In \emph{Proceedings of the 20th International Workshop on Semantic Evaluation ({S}em{E}val-2026)}.

\bibitem[{Singh and Das(2026)}]{culturag_blend2026}
Aditya Singh and Rickarya Das. 2026.
\newblock {C}ult{R}{A}{G} at {S}em{E}val-2026 {T}ask 7: {H}ybrid {S}parse-{D}ense {R}etrieval with {E}ntity-{C}entric {K}nowledge {B}ases for {C}ultural {M}{C}{Q} {A}nswering.
\newblock In \emph{Proceedings of the 20th International Workshop on Semantic Evaluation ({S}em{E}val-2026)}.

\bibitem[{Son et~al.(2024)Son, Lee, Kim, Kim, cheol Lee, Yeom, Jung, woo Kim, and Kim}]{son2024hae}
Guijin Son, Hanwool Lee, Suwan Kim, Huiseo Kim, Jae cheol Lee, Je~Won Yeom, Jihyu Jung, Jung woo Kim, and Songseong Kim. 2024.
\newblock Hae-rae bench: Evaluation of korean knowledge in language models.
\newblock In \emph{Proceedings of the 2024 Joint International Conference on Computational Linguistics, Language Resources and Evaluation (LREC-COLING 2024)}, pages 7993--8007.

\bibitem[{Song et~al.(2026)Song, Yeom, and Kim}]{knlpers_blend2026}
Jiwoo Song, Sihyeong Yeom, and Harksoo Kim. 2026.
\newblock {K}-{N}{L}{P}ers at {S}em{E}val-2026 {T}ask 7: {M}ultiple {L}{L}{M} {A}gent {D}ebate {S}ystem for {E}veryday {K}nowledge {A}cross {D}iverse {L}anguages and {C}ultures.
\newblock In \emph{Proceedings of the 20th International Workshop on Semantic Evaluation ({S}em{E}val-2026)}.

\bibitem[{Sriram and Sekar(2026)}]{models_without_borders_blend2026}
Swetha~Krishna Sriram and Nirupama Sekar. 2026.
\newblock {M}odels {W}ithout {B}orders at {S}em{E}val-2026 {T}ask 7: {B}ridging {C}ultural {C}ontexts with {S}earch-{G}rounded {Q}{A}.
\newblock In \emph{Proceedings of the 20th International Workshop on Semantic Evaluation ({S}em{E}val-2026)}.

\bibitem[{Tang et~al.(2026)Tang, Meng, and Jin}]{chengtang_blend2026}
Cheng Tang, Zhichao Meng, and Meizhi Jin. 2026.
\newblock chengtang at {S}em{E}val-2026 {T}ask 7: {A} {R}etrieval-{A}ugmented {G}eneration {F}ramework for {C}ultural {P}erspective {A}lignment in {E}veryday {M}{C}{Q}s.
\newblock In \emph{Proceedings of the 20th International Workshop on Semantic Evaluation ({S}em{E}val-2026)}.

\bibitem[{Tekanlou et~al.(2026)Tekanlou, Bakhtiyarzadeh, and Razmara}]{simorgh_blend2026}
Hadi Bayrami~Asl Tekanlou, Mahdi Bakhtiyarzadeh, and Jafar Razmara. 2026.
\newblock {S}imorgh at {S}em{E}val-2026 task 7: {R}egion-{A}ware {H}ybrid {R}etrieval for {L}ow-{R}esource {C}ultural {R}easoning in {M}ultilingual {Q}uestion {A}nswering.
\newblock In \emph{Proceedings of the 20th International Workshop on Semantic Evaluation ({S}em{E}val-2026)}.

\bibitem[{Wang et~al.(2026)Wang, Zhang, and Tan}]{wangkongqiang_blend2026}
Kongqiang Wang, Peng Zhang, and Qingli Tan. 2026.
\newblock Wangkongqiang at {S}em{E}val-2026 {T}ask 7: {E}veryday {K}nowledge {A}cross {D}iverse {L}anguages and {C}ultures.
\newblock In \emph{Proceedings of the 20th International Workshop on Semantic Evaluation ({S}em{E}val-2026)}.

\bibitem[{Yam and Yam(2026)}]{pinetree_blend2026}
Yen~Yee Yam and Hong~Meng Yam. 2026.
\newblock {P}inetree at {S}em{E}val-2026 {T}ask 7: {A} {L}arge-{S}cale {F}ailure {A}nalysis of {C}ultural {G}rounding in {L}anguage {M}odels.
\newblock In \emph{Proceedings of the 20th International Workshop on Semantic Evaluation ({S}em{E}val-2026)}.

\bibitem[{Yao and Yang(2026)}]{agentic_blend2026}
Xiao Yao and Liang Yang. 2026.
\newblock {A}gentic at {S}em{E}val-2026 {T}ask 7: {A} {Q}wen-based {S}ystem for {A}ccurate {R}etrieval of {E}veryday {K}nowledge {A}cross {D}iverse {L}anguages and {C}ultures.
\newblock In \emph{Proceedings of the 20th International Workshop on Semantic Evaluation ({S}em{E}val-2026)}.

\bibitem[{Zhou et~al.(2025)Zhou, Bamman, and Bleaman}]{zhou2025culture}
Naitian Zhou, David Bamman, and Isaac~L Bleaman. 2025.
\newblock Culture is not trivia: Sociocultural theory for cultural nlp.
\newblock In \emph{Proceedings of the 63rd Annual Meeting of the Association for Computational Linguistics (Volume 1: Long Papers)}, pages 25869--25886.

\bibitem[{Zhou et~al.(2024)Zhou, Bollegala, and Camacho-Collados}]{zhou-etal-2024-evaluating-short}
Yi~Zhou, Danushka Bollegala, and Jose Camacho-Collados. 2024.
\newblock \href {https://doi.org/10.18653/v1/2024.emnlp-main.1098} {Evaluating short-term temporal fluctuations of social biases in social media data and masked language models}.
\newblock In \emph{Proceedings of the 2024 Conference on Empirical Methods in Natural Language Processing}, pages 19693--19708, Miami, Florida, USA. Association for Computational Linguistics.

\bibitem[{Zhou et~al.(2023)Zhou, Camacho-Collados, and Bollegala}]{zhou-etal-2023-predictive}
Yi~Zhou, Jose Camacho-Collados, and Danushka Bollegala. 2023.
\newblock \href {https://doi.org/10.18653/v1/2023.emnlp-main.683} {A predictive factor analysis of social biases and task-performance in pretrained masked language models}.
\newblock In \emph{Proceedings of the 2023 Conference on Empirical Methods in Natural Language Processing}, pages 11082--11100, Singapore. Association for Computational Linguistics.

\end{thebibliography}
\newpage
\appendix
\section*{Appendix}

\begin{table}[!h]
\centering
\small
\begin{tabular}{ll}
\toprule
\textbf{Team} & \textbf{Citation} \\
\midrule
king001 & \cite{king001_blend2026}\\
wangkongqiang & \cite{wangkongqiang_blend2026} \\
Simorgh & \cite{simorgh_blend2026} \\
FLANS & \cite{flans_blend2026} \\
VerbaNexAI & \cite{verbanexAI_blend2026} \\
LocuPrompt & \cite{locuprompt_blend2026} \\
chengtang & \cite{chengtang_blend2026} \\
Agentic & \cite{agentic_blend2026} \\
HausaNLP & \cite{hausanlp_blend2026} \\
DFKI-MLT & \cite{dfki_mlt_blend2026} \\
Howard University & \cite{howard_uni_blend2026} \\
UTD-HLTRI & \cite{utd_hltri_blend2026} \\
CultRAG & \cite{culturag_blend2026} \\
uir-cis-7 & \cite{uircis7_blend2026} \\
GUIR & \cite{guir_blend2026} \\
Pinetree & \cite{pinetree_blend2026} \\
K-NLPers & \cite{knlpers_blend2026} \\
Models Without Borders & \cite{models_without_borders_blend2026} \\
\bottomrule
\end{tabular}
\caption{Teams and corresponding citations for submitted systems across Track A (SAQ) and Track B (MCQ).} 
\end{table}
\begin{table*}[t]
\centering
\scriptsize

\begin{subtable}{\textwidth}
\centering
\begin{adjustbox}{width=\textwidth}
\begin{tabular}{l*{15}{S}}
\toprule
ID &
{am-ET} & {en-ET} & {ar-DZ} & {en-DZ} &
{as-AS} & {en-AS} & {az-AZ} & {en-AZ} &
{zh-CN} & {en-CN} & {en-GB} & {en-US} &
{el-GR} & {en-GR} & {ha-NG} \\
\midrule
king001 & 97.8 & 97.8 & 97 & 97 & 99.2 & 99.2 & 99.6 & 99.6 & 100 & 100 & 99 & 99.6 & 99.8 & 99.8 & 94.2 \\
K-NLPERS & 50.6 & 38.6 & 49.2 & 49.2 & 49.2 & 49 & 59.2 & 56.2 & 78 & 66 & 72.6 & 75.6 & 60 & 59.4 & 40.8 \\
GUIR & 25.2 & 37.4 & 47.8 & 53.8 & 40.4 & 49.8 & 67 & 60.2 & 71.4 & 62.2 & 73.8 & 75.8 & 64.8 & 55.6 & 36 \\
rustycoder & 27.4 & 34.6 & 51.4 & 54.8 & 34.4 & 47.6 & 55.6 & 56.8 & 79 & 60.2 & 75.6 & 73.2 & 64.4 & 54.8 & 26.4 \\
yiyiyi & 37.2 & 41.6 & 45.8 & 64.8 & 32.4 & 56.2 & 38.8 & 41.8 & 66.8 & 69.4 & 61.6 & 77.6 & 40 & 43.6 & 34.4 \\
fedikallel & 32.6 & 34.4 & 45 & 46.4 & 40.6 & 40.6 & 51 & 52 & 68.6 & 57.6 & 68.2 & 70.8 & 58.4 & 52.2 & 30 \\
wangkongqiang & 24.2 & 41 & 43.2 & 53.6 & 31 & 44.2 & 56.8 & 58.8 & 79 & 67.2 & 75.6 & 73.6 & 52.4 & 53.6 & 28.8 \\
FLANS & 9.8 & 27.4 & 18.6 & 31.2 & 9.4 & 27 & 20.2 & 36.2 & 0 & 43.4 & 42.8 & 43.6 & 18.2 & 37.4 & 4.2 \\
alexrobertson4 & 19.4 & 19.4 & 18.2 & 18.2 & 18.4 & 18.4 & 13.8 & 13.8 & 35.6 & 70.8 & 82.6 & 86.8 & 14 & 13.8 & 9.4 \\
yx-ym & 10.4 & {} & 25.2 & {} & 13.4 & {} & 20 & {} & 59.2 & {} & 60 & 62.4 & 17 & {} & 23.8 \\
abaruah & {} & {} & {} & {} & 22 & 47.6 & {} & {} & {} & {} & {} & {} & {} & {} & {} \\
HausaNLP & {} & {} & {} & {} & {} & {} & {} & {} & {} & {} & {} & {} & {} & {} & 36.4 \\
DFKI-MLT & {} & {} & {} & {} & {} & {} & {} & {} & {} & {} & {} & {} & {} & {} & {} \\
qinchihongye & 97 & 97 & 97 & 97 & 99.2 & 99.2 & 99.6 & 99.6 & 100 & 100 & 99 & 99.6 & 99.8 & 99.8 & 94.2 \\
\midrule[\heavyrulewidth]
gpt-4.1-2025-04-14 & 35.40 & 42.20 & 58.60 & 61.20 & 47.60 & 54.20 & 68.40 & 61.80 & 74.20 & 68.80 & 77.40 & 80.20 & 69.40 & 64.20 & 32.40 \\
Qwen3 & 24.60 & 40.60 & 50.00 & 51.20 & 39.40 & 49.60 & 57.20 & 56.80 & 76.00 & 65.00 & 73.40 & 78.40 & 59.20 & 56.80 & 10.20 \\
\bottomrule
\end{tabular}
\end{adjustbox}
\end{subtable}

\vspace{6pt}

\begin{subtable}{\textwidth}
\centering
\begin{adjustbox}{width=\textwidth}
\begin{tabular}{l*{16}{S}}
\toprule
ID &
{en-NG} & {id-ID} & {en-ID} & {ko-KP} &
{en-KP} & {ko-KR} & {en-KR} & {fa-IR} &
{en-IR} & {es-MX} & {en-MX} & {es-ES} &
{en-ES} & {su-JB} & {en-JB} & {overall} \\
\midrule
king001 & 93.6 & 100 & 99.8 & 98.6 & 98.8 & 99.6 & 99.6 & 99.2 & 99.2 & 99.8 & 99.8 & 99.6 & 99.8 & 96.2 & 96.2 & 78.77 \\
K-NLPERS & 35.4 & 73.2 & 58.4 & 49.2 & 44.4 & 78 & 57.6 & 54 & 53.8 & 59.6 & 52.8 & 69.6 & 59.6 & 50.8 & 42.8 & 55.75 \\
GUIR & 32.6 & 73.2 & 57.8 & 46.6 & 46.2 & 76.4 & 65.8 & 64 & 58.6 & 69.6 & 58 & 74.8 & 59.4 & 49.2 & 45.4 & 55.50 \\
rustycoder & 31.8 & 72.6 & 63 & 43.6 & 44.8 & 70 & 58 & 64.6 & 55.2 & 64.8 & 54.4 & 70.4 & 58.8 & 38.4 & 45.4 & 52.74 \\
yiyiyi & 31.8 & 76.6 & 72.6 & 48.4 & 53.6 & 72.8 & 73.4 & 71.2 & 65.8 & 60.2 & 55.2 & 71.8 & 65.2 & 57.2 & 56 & 52.14 \\
fedikallel & 28.6 & 67 & 49.8 & 43.2 & 41 & 72.4 & 57.8 & 64 & 52.8 & 64 & 48.4 & 67.2 & 51 & 43.2 & 36.4 & 49.93 \\
wangkongqiang & 33.4 & 77 & 42.4 & 39 & 43.2 & 71.6 & 59.8 & 55.4 & 53 & 68.4 & 48 & 71.2 & 53 & 32.8 & 47.6 & 51.47 \\
FLANS & 21.8 & 35.8 & 43.2 & 29.2 & 12.4 & 41.8 & 41.4 & 20 & 38.2 & 28.2 & 40.8 & 33 & 40.8 & 7.8 & 23.6 & 26.02 \\
alexrobertson4 & 9.2 & 7.2 & 7.2 & 20.2 & 20.2 & 7.8 & 7.8 & 11.4 & 11.4 & 8.4 & 8.4 & 62.2 & 73.8 & 11.2 & 11.2 & 20.74 \\
yx-ym & {} & 44.4 & {} & 17.6 & {} & 19 & {} & 26.4 & {} & 47.8 & {} & 47.8 & {} & 16 & {} & {} \\
abaruah & {} & {} & {} & {} & {} & {} & {} & {} & {} & {} & {} & {} & {} & {} & {} & 34.80 \\
HausaNLP & {} & {} & {} & {} & {} & {} & {} & {} & {} & {} & {} & {} & {} & {} & {} & 36.40 \\
DFKI-MLT & {} & {} & {} & {} & {} & {} & {} & {} & {} & {} & {} & {} & {} & {} & {} & {} \\
qinchihongye & 93.6 & 100 & 99.8 & 98.6 & 98.8 & 99.6 & 99.6 & 99.2 & 99.2 & 99.8 & 99.8 & 99.6 & 99.8 & 96.2 & 96.2 & 78.10 \\
\midrule[\heavyrulewidth]
gpt-4.1 & 37.00 & 78.00 & 67.40 & 49.20 & 49.60 & 78.60 & 69.60 & 72.20 & 60.60 & 54.00 & 62.40 & 57.80 & 66.20 & 43.40 & 57.20 & 59.97 \\
Qwen3 & 32.40 & 71.40 & 64.40 & 40.60 & 38.60 & 64.60 & 60.00 & 59.40 & 54.20 & 63.40 & 60.00 & 70.40 & 61.20 & 37.20 & 47.40 & 53.79 \\
\bottomrule
\end{tabular}
\end{adjustbox}
\end{subtable}

\caption{SAQ results on the first version of \ours{}. }
\end{table*}
\begin{table*}[t]
\centering
\scriptsize

\begin{subtable}{\textwidth}
\centering
\begin{adjustbox}{width=\textwidth}
\begin{tabular}{l*{8}{S}}
\toprule
ID &
{am-ET} & {ar-DZ} & {as-AS} & {az-AZ} &
{el-GR} & {en-GB} & {en-US} & {es-ES} \\
\midrule
GUIR & 96.54 & 99.38 & 99.51 & 98.91 & 98.35 & 99.17 & 98.56 & 99.12 \\
uir-cis-7 & 98.67 & 98.15 & 99.35 & 99.22 & 98.1 & 98.25 & 98.66 & 98.55 \\
Pinetree & 69.96 & 92.58 & 81.15 & 87.85 & 94.37 & 96.59 & 94.23 & 94.3 \\
K-NLPers & 66.36 & 92.23 & 78.95 & 81.28 & 93.93 & 95.8 & 95.62 & 95.18 \\
DFKI-MLT & 58.85 & 93.27 & 80.5 & 82.06 & 88.44 & 96.12 & 94.13 & 93.73 \\
Models without Borders & 69.26 & 86.46 & 66.71 & 62.52 & 81.27 & 81.26 & 88.98 & 85.19 \\
Agentic & 58.02 & 75.15 & 70.18 & 74.36 & 81.6 & 86.25 & 90.47 & 88.24 \\
LocuPrompt & 57.21 & 81.31 & 65.85 & 63.43 & 84.09 & 82.1 & 90.63 & 84 \\
Simorgh & 54.91 & 80.35 & 71.03 & 70.48 & 76.88 & 87.49 & 84.96 & 82.03 \\
FLANS & {} & {} & {} & {} & {} & 92.2 & 88.31 & 83.79 \\
\midrule[\heavyrulewidth]
gpt-4.1 & 77.16 & 88.12 & 83.39 & 79.89 & 92.17 & 96.82 & 95.21 & 95.39 \\
Qwen3 & 68.04 & 90.77 & 73.64 & 80.80 & 93.93 & 96.22 & 92.69 & 94.82 \\
\bottomrule
\end{tabular}
\end{adjustbox}
\end{subtable}

\vspace{6pt}

\begin{subtable}{\textwidth}
\centering
\begin{adjustbox}{width=\textwidth}
\begin{tabular}{l*{9}{S}}
\toprule
ID &
{es-MX} & {fa-IR} & {ha-NG} & {id-ID} &
{ko-KP} & {ko-KR} & {su-JB} & {zh-CN} & {Overall} \\
\midrule
GUIR & 99.32 & 98.81 & 98.11 & 99.85 & 91.9 & 98.96 & 100 & 98.51 & 98.51 \\
uir-cis-7 & 98 & 97.95 & 96.07 & 97.59 & 94.6 & 99.76 & 97.82 & 98.16 & 98.16 \\
Pinetree & 94.1 & 81.32 & 76.94 & 88.42 & 80.64 & 93.27 & 93.36 & 88.30 & 88.30 \\
K-NLPers & 94.1 & 79.1 & 79.43 & 87.77 & 73.5 & 90.13 & 90.88 & 86.36 & 86.36 \\
DFKI-MLT & 94.94 & 78.56 & 74.35 & 87.02 & 78.31 & 86.86 & 92.79 & 85.02 & 85.02 \\
Models without Borders & 89.2 & 70.18 & 64.24 & 72.63 & 65.81 & 87.34 & 74.65 & 75.82 & 75.82 \\
Agentic & 92.84 & 64.1 & 62.3 & 75.49 & 58.4 & 87.3 & 79.26 & 75.13 & 75.13 \\
LocuPrompt & 89.36 & 62.34 & 58.81 & 72.08 & 55.51 & 86.07 & 87.2 & 74.43 & 74.43 \\
Simorgh & 87.41 & 64.23 & 60.11 & 74.49 & 64.3 & 80.65 & 87.66 & 74.13 & 74.13 \\
FLANS & 82.2 & {} & {} & {} & {} & {} & 80.82 & 85.46 & 85.46 \\
Verbanex-AI & 15.6 & 16.2 & 31.2 & 24.2 & 21.2 & 12.8 & 13.8 & 17.8 & 17.80 \\
UTD-HLT & {} & {} & {} & {} & {} & {} & {} & {} & {} \\
aditya26189 & 94 & 73.02 & 69.52 & {} & 66.18 & 83.4 & 80.72 & 77.89 & 77.89 \\
narjes & {} & {} & {} & {} & {} & {} & {} & 91.79 & 91.79 \\
\midrule[\heavyrulewidth]
gpt-4.1 & 94.79 & 79.97 & 78.54 & 87.07 & 82.75 & 91.64 & 93.11 & 87.85 & 87.85 \\
Qwen3 & 92.58 & 78.21 & 75.90 & 87.87 & 74.32 & 90.80 & 92.59 & 85.12 & 85.12 \\
\bottomrule
\end{tabular}
\end{adjustbox}
\end{subtable}

\caption{MCQ results on the first version of \ours{}.}
\end{table*}

\onecolumn

\end{document}